\pdfoutput=1
\documentclass[11pt]{article}

\usepackage[final]{acl}

\usepackage{times}
\usepackage{latexsym}

\usepackage[T1]{fontenc}
\usepackage[utf8]{inputenc}

\usepackage{microtype}

\usepackage{inconsolata}

\usepackage{graphicx}
\usepackage{amssymb}
\usepackage{amsmath}
\usepackage{booktabs}
\usepackage{microtype}      % microtypography
\usepackage{xcolor}         % colors
\usepackage{graphicx} % For including images
\usepackage{multirow}
\usepackage{pgfplots}
\pgfplotsset{compat=1.18}
\usepackage{pgfplotstable}
\usepackage{tikz}
\usepackage{subcaption}
\usepackage{xspace}
\def\secref#1{\S\ref{sec:#1}}
\def\seclabel#1{\label{sec:#1}}

\def\ofa{\textsc{Ofa}\xspace}
\def\frameworkname{\textsc{HyperOfa}\xspace}

\title{\textsc{HyperOfa}: Expanding LLM Vocabulary to New
Languages via Hypernetwork-Based Embedding
Initialization}

\author{
 \textbf{Enes Özeren\textsuperscript{*}},
 \textbf{Yihong Liu\textsuperscript{*\(\diamond\)}},
 \textbf{Hinrich Schütze\textsuperscript{*\(\diamond\)}}
\\
\\
 \textsuperscript{*}LMU Munich \,\,\,\,
 \textsuperscript{\(\diamond\)}Munich Center for Machine Learning (MCML)
\\
\texttt{enes.oezeren@campus.lmu.de}
}

\begin{document}
\maketitle
\begin{abstract}
% yihong's write
Many pre-trained language models (PLMs) exhibit suboptimal performance on mid- and low-resource languages, largely due to limited exposure to these languages during pre-training.
A common strategy to address this is to introduce new tokens specific to the \emph{target} languages, initialize their embeddings, and apply \emph{continual pre-training} on target-language data.
Among such methods, \ofa \citep{liu-etal-2024-ofa} proposes a similarity-based subword embedding initialization heuristic that is both effective and efficient. However, \ofa restricts target-language token embeddings to be convex combinations of a fixed number of source-language embeddings, which may limit expressiveness.
To overcome this limitation, we propose \frameworkname, a hypernetwork-based approach for more adaptive token embedding initialization. 
The hypernetwork is trained to map from an \emph{external multilingual word vector space} to the \emph{PLM’s token embedding space} using source-language tokens.\footnote{We will use \emph{vector space} and \emph{embedding space} to refer to the two different spaces for convenience.}
Once trained, it can generate flexible embeddings for target-language tokens, serving as a good starting point for continual pretraining.
Experiments demonstrate that \frameworkname consistently outperforms random initialization baseline and matches or exceeds the performance of \ofa in both continual pre-training convergence and downstream task performance. 
We make the code publicly available.\footnote{\url{https://github.com/enesozeren/hyper-ofa}}

% Most pre-trained language models are trained primarily for high-resource languages, limiting their usability in mid- and low-resource languages. A common approach to adapt these models for other languages involves introducing new tokens specific to these languages and continuing pre-training. 
% However, the method used to initialize these newly introduced tokens significantly impacts the efficiency of continued pre-training. Poor initialization can lead to longer training times and increased computational costs. 
% OFA \citep{liu-etal-2024-ofa} provides an effective initialization strategy based on a similarity-driven heuristic. However, it relies on existing token embeddings to initialize new ones, uses linear combination methods and some hyper-parameters which may limit flexibility. To overcome this, we propose HyperOFA, a hypernetwork-based approach designed to generate more adaptive token initializations. Experimental results show that HyperOFA consistently outperforms the random initialization baseline and performs on par with OFA, in some cases surpassing it.
\end{abstract}

\section{Introduction}

Multilingual PLMs,  trained on massive multilingual corpora, have achieved impressive performance across many high-resource languages \citep{devlin-etal-2019-bert,artetxe-etal-2020-cross,liang-etal-2023-xlm,ustun-etal-2024-aya}.
However, such models often perform suboptimally on languages that are under-resourced in their pre-training data \citep{wu-dredze-2020-languages}, and in extreme cases, they perform poorly on entirely unseen languages \citep{adelani-etal-2024-sib}, particularly when there is minimal lexical overlap or shared vocabulary
between these unseen languages and the languages covered by the PLM \citep{muller-etal-2021-unseen,moosa-etal-2023-transliteration,liu-etal-2024-translico,xhelili-etal-2024-breaking}. 

A common strategy for adapting PLMs to such under-resourced or unseen languages is to introduce new, language-specific tokens, initialize their embeddings, and continually pre-train the model on data from the target languages \citep{tran2020english}.\footnote{We simply use \emph{source tokens} to refer to tokens belonging to the source languages that are already covered in the PLM vocabulary. Similarly, \emph{target tokens} is used to refer to tokens that belong to the target languages that one wants to adapt to and are usually not covered by the PLM vocabulary.}
A key challenge in this process lies in the initialization of these new token embeddings.
A naive approach would be random initialization from a given simple distribution, e.g., multivariate Gaussian, \citep{hewitt2021initializing,de-vries-nissim-2021-good,marchisio-etal-2023-mini}. However, such an initialization fails to leverage any lexical or semantical knowledge captured by the original source-language embeddings.

To address this, recent work has explored more informed initialization strategies, using similarity-based heuristics to better align the initialized target embeddings with the existing embedding space, thereby enhancing language adaptation and accelerating continual pre-training \citep{minixhofer-etal-2022-wechsel,dobler-de-melo-2023-focus,liu-etal-2024-ofa,mundra-etal-2024-empirical,yamaguchi-etal-2024-empirical,yamaguchi2024can}. 
Among this line of work, for example, \ofa \citep{liu-etal-2024-ofa} uses external multilingual word vectors to compute similarities between source and target tokens, then initializes target embeddings as convex combinations of source embeddings, weighted by these similarities.
This approach ensures the target embeddings reside in the same vector space as the source ones.
However, the similarity-based convex combination restricts the relation between embeddings of source tokens and target tokens to be linear, which might not be expressive enough considering the non-linearity of Transformer \citep{vaswani2017attention}.

To overcome this limitation, this paper presents \frameworkname, a hypernetwork-based initialization method designed to enhance the expressiveness and adaptability of embedding initialization. Rather than depending on similarity heuristics, we explicitly learn a mapping from an external vector space to the PLM’s embedding space via a hypernetwork.
The hypernetwork is trained to predict the embedding of a source token, given external multilingual word vectors of a small set of related words as input.
Training proceeds by minimizing the discrepancy between the predicted and actual PLM embeddings of source tokens.
Once trained, the hypernetwork is used to generate embeddings for target tokens, providing a robust initialization for continual pre-training on the target languages.

To evaluate \frameworkname, we follow the experimental setup of \ofa, adapting both a monolingual PLM, i.e., RoBERTa \citep{liu2019roberta}, and a multilingual PLM, i.e., XLM-R \citep{conneau-etal-2020-unsupervised}, to 22 languages covering high-, mid-, and low-resource scenarios. 
We investigate two research questions: (\textbf{1}) \emph{How well do the initialized embeddings perform on their own?} 
and (\textbf{2}) 
\emph{How effective are they as a starting point for continual pre-training?}
To answer these, we evaluate models before and after continual pre-training via zero-shot cross-lingual transfer on downstream tasks, including sentence retrieval and sequence labeling.
Our empirical results show that \frameworkname consistently outperforms the random initialization and achieves competitive or superior performance compared to \ofa. Our contributions are as follows:

\begin{itemize}
    \item We propose \frameworkname, a hypernetwork-based method for initializing embeddings of new tokens in target languages.
    \item We extensively evaluate \frameworkname on adapting RoBERTa and XLM-R to many languages and various downstream tasks.
    \item We show that \frameworkname outperforms random initialization and matches or exceeds the performance of its counterpart \ofa.
\end{itemize}

\section{Related Work}

\paragraph{Tokenizer and Vocabulary Manipulation}

Manipulating an existing PLM's vocabulary and its accompanying tokenizer is a common approach for adapting it to new languages \citep{pfeiffer-etal-2021-unks,alabi-etal-2022-adapting,zeng2023greenplm,cui2024efficient} or new domains \citep{lamproudis2022vocabulary,liu-etal-2023-task,Balde2024vocabulary}. Typically, another tokenizer is trained on the target data using the same tokenization algorithm as used by the original one, such as Byte-Pair Encoding \citep{gage1994new,sennrich-etal-2016-neural}, WordPiece \citep{Schuster2012wordpiece,wu2016google}, and SentencePiece \citep{kudo-richardson-2018-sentencepiece,kudo-2018-subword}. Then, the new tokenizer is merged with the original tokenizer, where unseen tokens are added, resulting in a large vocabulary. \citet{imanigooghari-etal-2023-glot500} successfully apply such a pipeline to extend the language coverage of XLM-R \citep{conneau-etal-2020-unsupervised} to more than 500 languages. Similarly, \citet{liu-etal-2025-transmi} adapts XLM-R to transliterated data by merging romanized subwords into the vocabulary.

\paragraph{Target Embedding Initialization}

The embeddings for the new tokens have to be initialized before the model can be used or continually pre-trained. The simplest approach is to randomly initialize the new token embeddings \citep{artetxe-etal-2020-cross,de-vries-nissim-2021-good,alabi-etal-2022-adapting,imanigooghari-etal-2023-glot500}. To better leverage the already encoded knowledge in the PLM, some work tries to initialize the new target token embeddings as linear combinations of embeddings of the source tokens, weighted by similarities between target and source tokens. An early work, \citet{tran2020english}, induces such similarities from a parallel corpus. More recently, another line of work explores the possibility of directly inducing such similarities from well-aligned external word embeddings \citep{minixhofer-etal-2022-wechsel, dobler-de-melo-2023-focus, liu-etal-2024-ofa,yamaguchi-etal-2024-empirical, yamaguchi2024can,ye-etal-2024-mosecrot}. However, the similarity-based convex combination might restrict the expressiveness of the new token embeddings. Therefore, this work aims to improve the initialization by breaking the linearity obstacle.

\paragraph{Hypernetworks}

Hypernetworks are neural networks designed to generate the weights of another network \citep{ha2016hypernetworks, chauhan2024brief}. A recent survey by \citet{chauhan2024brief} highlights their application across various domains such as computer vision \citep{Oswald2020Continual} and natural language processing (NLP) \citep{volk-etal-2023-example, pinter-etal-2017-mimicking, schick-schutze-2020-bertram, minixhofer2024zero}. One of the earlier works in initializing embeddings with hypernetworks is MIMICK \citep{pinter-etal-2017-mimicking}, which focuses on predicting the out-of-vocabulary word embeddings with a hypernetwork. Similarly, \citet{schick-schutze-2020-bertram} integrates a hypernetwork into BERT \citep{devlin-etal-2019-bert} to generate embeddings for rare words. More recently, \citet{minixhofer2024zero} proposed a hypernetwork-based method for zero-shot tokenizer transfer, enabling a language model to detach from its tokenizer. Our work builds upon the insights from this line of work and designs a hypernetwork to map from the external word vector space to the PLM's embedding space, allowing for wise initialization of the new token embeddings for effective continual pre-training. 

% Similar to WECHSEL and \ofa methods, \frameworkname also utilizes an external source of information. However, the key distinction of \frameworkname lies in its use of a hypernetwork to directly predict embeddings for new token initialization, whereas the previous two methods initialize new tokens by combining embeddings of existing ones with a linear combination technique. \frameworkname offers advantages for initializing tokens from languages semantically distant from the model's original ones, leveraging a hypernetwork to capture non-linear relationships during this process.

\section{Methodology}
\seclabel{sec:methodology}

\begin{figure*}[h!] % Optional: [h!] places the figure "here" if possible
   \centering
  \includegraphics[width=0.8\linewidth]{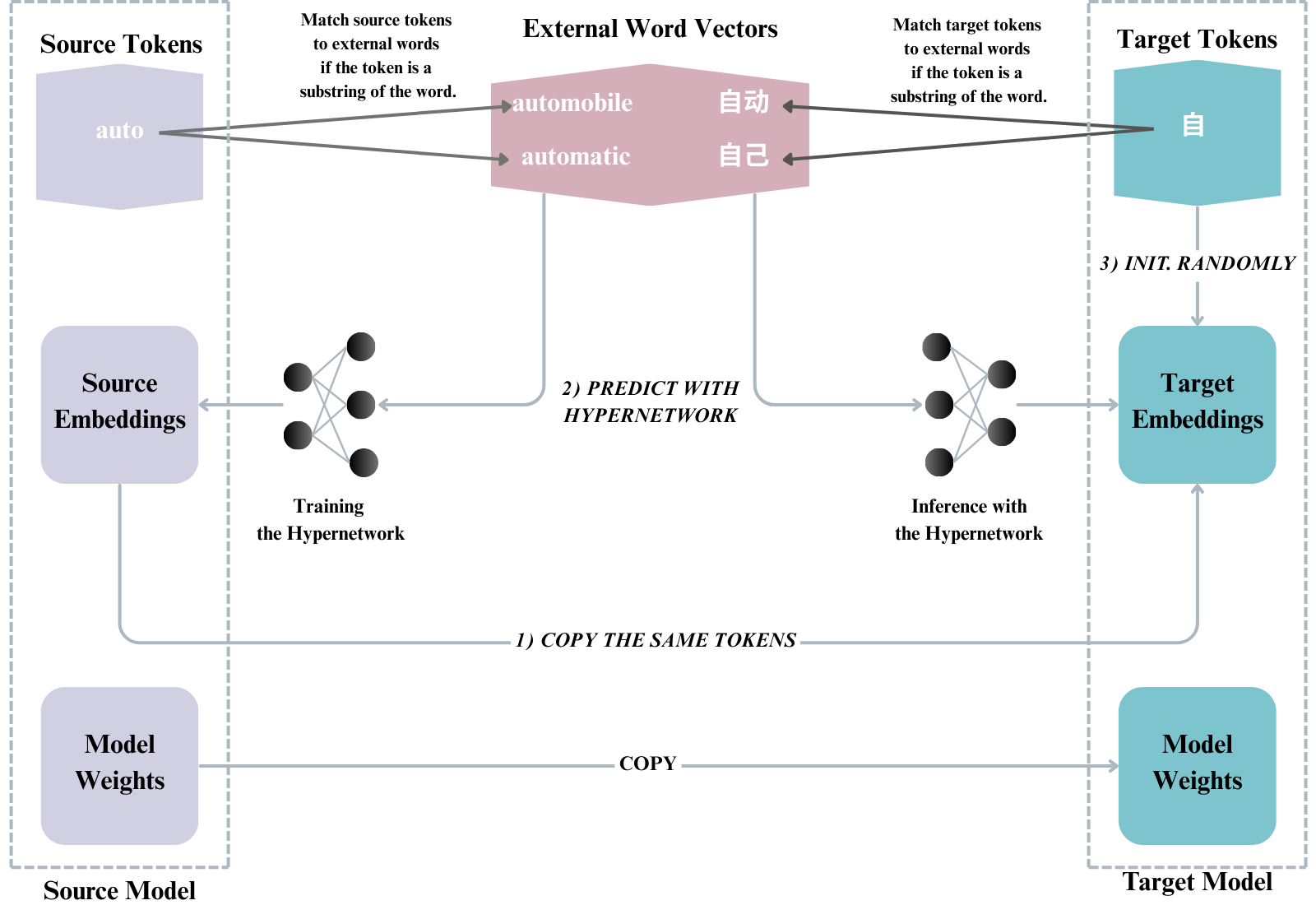} % Adjust the width as needed
  \caption{\frameworkname pipeline. The source model (left) transfers weights to the target model (right). The target embeddings are initialized by first copying embeddings for matching tokens, then generating embeddings via a hypernetwork for tokens with matching external words, and finally randomly initializing the rest.}
  \label{fig:hyperofa_diagram}
\end{figure*}

\frameworkname builds upon certain aspects of \ofa \citep{liu-etal-2024-ofa}, e.g., factorized parameterization (cf. \secref{sec:factorization}) and external multilingual vector vectors (cf. \secref{sec:external_word_vectors}). The key differentiator is that we directly predict the target token embeddings using a hypernetwork (cf. \secref{sec:hypernetwork}) instead of initialization based on similarity-heuristics. For a clearer understanding, we therefore follow the notations used by \citet{liu-etal-2024-ofa} and introduce \frameworkname in the following. Figure~\ref{fig:hyperofa_diagram} provides an overview of \frameworkname.

\subsection{Problem Setting}
\seclabel{sec:problem_setting}

Given a model with a source tokenizer $TOK^s$ with vocabulary $V^s$, the goal is to replace the source tokenizer with a target tokenizer $TOK^t$ with vocabulary $V^t$ that supports a broader range of tokens across various languages. 
Typically, $|V^s| < |V^t|$.
The core problem is to \textbf{initialize the target embeddings $\boldsymbol{E}^t \in \mathbb{R}^{|V^t| \times D}$}, where $D$ is the embedding dimension, which is the same as the dimension of the source embeddings $\boldsymbol{E}^s \in \mathbb{R}^{|V^s| \times D}$.
% Starting from the source embeddings $\boldsymbol{E}^s \in \mathbb{R}^{|V^s| \times D}$, the \frameworkname method aims to initialize target embeddings $\boldsymbol{E}^t \in \mathbb{R}^{|V^t| \times D}$ wisely by keeping the same embedding dimension $D$.

\subsection{Source Embedding Factorization}
\seclabel{sec:factorization}

Since $|V^t| > |V^s|$, the number of embedding parameters grows significantly from $V^s \times D$ to $V^t \times D$ in the target model.
This can result in a large ratio of model parameters in the embedding matrix, limiting the efficiency. 
To address this, \citet{liu-etal-2024-ofa} adopts a factorized parametrization to represent the embeddings, similar to \citet{lan2019albert}.

Factorization decomposes the $\boldsymbol{E}^s$ into two smaller matrices using the Singular Value Decomposition (SVD) method, such that $\boldsymbol{E}^s \approx \boldsymbol{F^s}\boldsymbol{P}$, where $\boldsymbol{F}^s \in \mathbb{R}^{|V^s| \times D'}$ is the coordinate matrix containing token-specific parameters, and $\boldsymbol{P} \in \mathbb{R}^{D' \times D}$ is the primitive embedding matrix capturing language-agnostic features. 
When $D' < D$, the total number of parameters of $\boldsymbol{F}^s$ and $\boldsymbol{P}$ is smaller than $\boldsymbol{E}^s$. 
Since $\boldsymbol{P}$ is expected to be shared across languages, one only needs to initialize the coordinate matrix $\boldsymbol{F}^t \in \mathbb{R}^{|V^t| \times D'}$ for $TOK^t$ while reusing the same $\boldsymbol{P}$.
The original dimension can be restored by multiplication: $\boldsymbol{F}^tP \in \mathbb{R}^{|V^t| \times D}$.

% \citet{liu-etal-2024-ofa} experiments with different $D'$ values to explore the trade-off between parameter saving and model performance.

\subsection{Matching External Word Vectors}
\seclabel{sec:external_word_vectors}

\ofa \citep{liu-etal-2024-ofa} takes advantage of external well-aligned multilingual vectors $W$ to induce the similarities between source tokens and target tokens.\footnote{\citet{liu-etal-2024-ofa} use $\overset{\longrightarrow}{\text{ColexNet+}}$ \citep{liu-etal-2023-crosslingual-transfer}, which are static word vectors that contain over 4M words spanning more than 1K languages. The tokens in $V^t$ are usually subwords of the word types covered by $\overset{\longrightarrow}{\text{ColexNet+}}$.}
In contrast, we directly use these vectors to train a hypernetwork to map from the vector space to the embedding space, discarding the similarity-based heuristics.
To do this, we first need to create corresponding pairs of tokens in $V^s \cup V^t$ and words in $W$, which is achieved by a matching operation. 
Specifically, a token in $V^s \cup V^t$ is matched with a word in $W$ if that word contains the token as a subword (cf. Figure~\ref{fig:hyperofa_diagram}).
This matching operation results in ${s_i}$ (resp. ${t_j}$), a set of matched words for each token $i$ in $V^s$ (resp. each token $j$ in $V^t$).
We then represent the set of matched word vectors for each token $i$ (resp. $j$) as $W_{\{s_i\}}$ (resp. $W_{\{t_j\}}$).

\subsection{Hypernetwork}
\seclabel{sec:hypernetwork}

To address the main limitation of \ofa -- use a convex combination of source-token embeddings to initialize the target embeddings -- we propose a hypernetwork approach to directly map from the vector space to the embedding space, which introduces non-linearity, and thus is more expressive.

After performing factorization (cf. \secref{sec:factorization}) and creating the set of matched words and tokens (cf. \secref{sec:external_word_vectors}), a hypernetwork $HN_\theta$ with parameters $\theta$ is introduced.
The ultimate aim of the hypernetwork is to generate the target-token embedding $\boldsymbol{F}_j$ by using the matched word vectors $W_{\{t_j\}}$, where $j \in V^t$.
Therefore, we need to properly train $HN_\theta$ so that it can map from the vector space to the embedding space.
To do this, we create a training set for $HN_\theta$. Each item in the training set is a pair: ($W_{\{s_i\}}$, $\boldsymbol{F}^s_i$), where $W_{\{s_i\}}$ and $\boldsymbol{F}^s_i$ are the set of matched word vectors and coordinate vector in $\boldsymbol{F}^s$ for token $i$ in $V^s$, respectively.\footnote{ We exclude ($W_{\{s_i\}}$, $\boldsymbol{F}^s_i$) from the training set if ${s_i} = \emptyset$, i.e., there are no matched words for the concerned token $i$.}
$HN_\theta$ then takes $W_{\{s_i\}}$ as input and is trained to predict $\boldsymbol{F}^s_i$.

A custom loss function is proposed for the training, which contains two training objectives: a batch-wise \emph{contrastive loss} $\mathcal{L}_{\text{c}}$ and a \emph{normalized L1 loss} $\mathcal{L}_{\text{L1}}$.
The contrastive loss $\mathcal{L}_{\text{c}}$ aims to improve the similarity between the ground-truth coordinate embeddings and the predictions:

$$
\mathcal{L}_{\text{c}} = \mathbb{E}\left[-{\log \frac{\exp(\text{sim}(\boldsymbol{F}_i^s, \hat{\boldsymbol{F}}_i^s) / \tau)}{{ \exp(\text{sim}(\boldsymbol{F}_i^s, \hat{\boldsymbol{F}}_i^s) / \tau)}+\text{NEG}}}\right]
$$
where  $\text{NEG} = \sum_{k \neq i}\exp(\text{sim}(\boldsymbol{F}_k^s, \hat{\boldsymbol{F}}_i^s))/\tau)$, $\text{sim}$ is cosine similarity, $\hat{\boldsymbol{F}}_i^s = HN_\theta(W_{\{s_i\}})$ and $\tau$ is temperature. 
The normalized L1 loss $\mathcal{L}_{\text{L1}}$ aims to preserve magnitude consistency:
$$
\mathcal{L}_{\text{L1}} = \mathbb{E}\left[{\|\boldsymbol{F}_i^s - \hat{\boldsymbol{F}}_i^s\|_1}\right]
$$
The final loss is 
$
    \mathcal{L}(\theta) = \lambda \cdot \mathcal{L}_{\text{c}} + (1 - \lambda) \cdot \mathcal{L}_{\text{L1}}
$ where $\lambda$ is a hyperparameter controlling the weight. 

% here

% During training, a custom loss function is employed to ensure that the predicted token embeddings not only align semantically with the target embeddings but also maintain a similar norm. This loss combines contrastive loss, which enforces semantic similarity, and normalized L1 loss, which preserves magnitude consistency.

% \begin{equation}
%     \mathcal{L}(\theta) = \lambda \cdot \mathcal{L}_{\text{contr.}} + (1 - \lambda) \cdot \mathcal{L}_{\text{L1}}
% \end{equation}

% \small 
% \begin{align}
% \mathcal{L}_{\text{contr.}} = -\frac{1}{N} \sum_{k=1}^{N}{\log \frac{\exp(\text{sim}(F_k^s, \hat{F}_k^s) / T)}{\sum_{l=1}^{N}{ \exp(\text{sim}(F_l^s, \hat{F}_l^s) / T)}}} \\
% \mathcal{L}_{\text{L1}} = \frac{1}{N} \sum_{k=1}^{N}{\frac{\|F_k^s - HN_\theta(W(s_k))\|_1}{-\frac{1}{N} \sum_{l=1}^{N}{\|F_l^s\|}}}
% \end{align}

% \normalsize % Reset to normal size after equations

% Here the $N$ is batch size, $\text{sim}$ is cosine similarity, $T$ is temperature and $\lambda$ is a parameter to balance loss components. The $\mathcal{L}_{\text{L1}}$ component is a normalized by average embedding value to prevent very small loss values.

When designing the model architecture for $HN_\theta$, there are certain requirements because of the input -- a set of vectors. 
First, the number of matched word vectors may vary for different tokens, meaning the model architecture must be capable of handling variable-length inputs. 
Secondly, since the order of the input matched word vectors should not influence the prediction, the model should be permutation-invariant.
Considering these requirements,
we used a BiLSTM \citep{schuster1997bidirectional} for $HN_\theta$ despite it not inherently satisfying the permutation-invariance requirement.\footnote{We experimented with both Transformer and BiLSTM architectures for the hypernetwork, but experiments have shown that BiLSTM works better in our study (cf. Appendix \secref{sec:hypernetwork_architecture})}
To address the BiLSTM's sensitivity to input order, data augmentation is implemented by randomly shuffling the order of the word vectors during each training epoch, effectively preventing the model from overfitting to specific sequence arrangements.

% The trained BiLSTM-based hypernetwork $HN_\theta$ generates target embeddings by taking matched word vectors $W(t_j)$ as inputs, with the network parameters optimized using the specified loss function.

\subsection{New Token Initialization}
\seclabel{sec:new_token_init}

The target coordinate embeddings, $\boldsymbol{F}^t$, are initialized in three steps similar to \ofa \citep{liu-etal-2024-ofa} (cf. Figure~\ref{fig:hyperofa_diagram}).

\begin{enumerate}
    \item For tokens in $V^s \cap V^t$, their embeddings in $\boldsymbol{F}^s$ are directly copied to $\boldsymbol{F}^t$. 
    \item For tokens that have at least one matched word (cf. \secref{sec:external_word_vectors}), their embeddings are predicted by $HN_\theta$ using the set of vectors $W_{\{t_j\}}$ as input.
    \item For the remaining tokens, their embeddings are randomly initialized from a normal distribution $\mathcal{N}(\mathbb{E}[F^s, \text{Var}[F^s])$, similar to \ofa.
\end{enumerate}

\section{Experimental Setup}

% In this study, the vocabularies of RoBERTa-base (RoBERTa) \citep{liu2019roberta} (50K tokens) and XLM-RoBERTa-base (XLM-R) \citep{conneau-etal-2020-unsupervised} (250K tokens) are extended to match the Glot500-m vocabulary \citep{imanigooghari-etal-2023-glot500}, which contains 401K tokens. 

% Before initializing new token embeddings, the factorization method described in \secref{sec:factorization} is applied for different dimensions $D'$. While the original embedding size for both models is 768, factorization is used to create coordinate matrices $\boldsymbol{F}^s$ with dimensions of 100, 200, and 400. The models with reduced embedding sizes are referred to as RoBERTa-100, RoBERTa-200, RoBERTa-400, and similarly, XLM-R-100, XLM-R-200, XLM-R-400.

% To compare the performance of \frameworkname with \ofa, different hypernetworks are trained to initialize the new token embeddings for these RoBERTa-xxx and XLM-R-xxx models.

\subsection{\frameworkname-Based Models}

Following \ofa \citep{liu-etal-2024-ofa}, we use the tokenizer of Glot500-m \citep{imanigooghari-etal-2023-glot500} as the target tokenizer, which is trained by SentencePiece \citep{kudo-richardson-2018-sentencepiece,kudo-2018-subword} and has a vocabulary size of 401K. We consider three different dimensions for $D'$: 100, 200, 400 (cf. \secref{sec:factorization}). We create 6 models using \frameworkname as follows:

\paragraph{\frameworkname-mono-xxx}
These are RoBERTa models \citep{liu2019roberta} with an extended vocabulary (from the original 50K to 401K). ``xxx'' denotes the embedding dimension of the model (100, 200, 400), and the "mono" suffix indicates that the model is originally monolingual. The new token embeddings are predicted by a hypernetwork trained specifically for each model (cf. \secref{sec:hypernetwork_setup}) or randomly initialized as a fallback (cf. \secref{sec:new_token_init}).

\paragraph{\frameworkname-multi-xxx}
These are XLM-R models \citep{conneau-etal-2020-unsupervised} with an extended vocabulary (from the original 250K to 401K). ``xxx'' denotes the embedding dimension of the model (100, 200, 400), and the "multi" suffix indicates that the model is originally multilingual. The new token embeddings are predicted by a hypernetwork trained specifically for each model (cf. \secref{sec:hypernetwork_setup}) or randomly initialized as a fallback (cf. \secref{sec:new_token_init}).

\subsection{Hypernetwork Setup}\seclabel{sec:hypernetwork_setup}

\paragraph{Hypernetwork Training Dataset} 

For \frameworkname-mono-xxx models, the hypernetwork training dataset consists of \textbf{22K pairs} of embeddings of the source tokens and their corresponding sets of matched word vectors, as 22K out of RoBERTa’s 50K vocabulary tokens match at least one word in $\overset{\longrightarrow}{\text{ColexNet+}}$ (cf. \secref{sec:hypernetwork}).
Similarly, for XLM-R, the training dataset contains \textbf{103K pairs}, corresponding to 103K tokens from its 250K vocabulary. 
% To mitigate overfitting, data augmentation is applied by shuffling word vector order before each epoch.
% Additionally, with 50\% probability, the number of word vectors is randomly limited to 50–100\% of the available vectors. 
% A custom sampler is also employed to ensure that batches contain input sequences of similar length, improving training stability.

\paragraph{Hypernetwork Training}

\begin{figure}[h!]
    \centering
    \begin{tikzpicture}
      \begin{axis}[
          width=0.48\textwidth,
          height=0.32\textwidth,
          xlabel={Epoch},
          ylabel={Validation Cosine Similarity},
          legend pos=south east,
          grid=both,
          axis lines=box,
          every axis/.append style={font=\small},          
          ymin=0,
          ymax=0.2,
          y tick label style={
                /pgf/number format/fixed,
                /pgf/number format/precision=3
            }
      ]
        % CSV1 data (insert your comma-separated values here)
        \addplot+[mark=*] table[
          x expr=\coordindex, 
          y=value,
          col sep=comma
        ] {
        value
        0.058
        0.061
        0.092
        0.098
        0.108
        0.108
        0.107
        0.12
        0.108
        0.122
        0.123
        0.128
        0.121
        0.105
        0.13
        0.132
        0.134
        0.136
        0.137
        0.137
        0.138
        0.138
        0.14
        0.143
        0.144
        0.145
        0.146
        0.146
        0.148
        0.15
        0.151
        0.153
        0.153
        0.156
        0.155
        0.158
        0.158
        0.16
        0.157
        0.162
        0.163
        0.163
        0.164
        0.165
        0.165
        0.167
        0.168
        0.169
        0.169
        0.171
        0.171
        0.17
        0.172
        0.172
        0.172
        0.172
        0.173
        0.174
        0.173
        0.175
        0.173
        0.176
        0.175
        0.175
        0.175
        0.175
        0.175
        0.175
        0.176
        0.175
        0.177
        0.177
        0.176
        0.176
        0.176
        0.177
        0.177
        0.177
        0.177
        0.177
        0.177
        0.177
        0.176
        0.176
        0.177
        0.177
        0.177
        0.177
        0.177
        0.177
        0.177
        0.177
        0.178
        0.176
        0.176
        0.177
        0.177
        0.176
        0.178
        0.177
        };
        \addlegendentry{Large (210M) BiLSTM Hypernetwork}

        % CSV2 data (insert your comma-separated values here)
        \addplot+[mark=square*] table[
          x expr=\coordindex, 
          y=value,
          col sep=comma
        ] {
        value
        0.045
        0.079
        0.09
        0.102
        0.108
        0.11
        0.113
        0.115
        0.116
        0.12
        0.121
        0.124
        0.124
        0.125
        0.123
        0.124
        0.128
        0.131
        0.132
        0.133
        0.136
        0.138
        0.139
        0.139
        0.138
        0.139
        0.141
        0.142
        0.142
        0.144
        0.145
        0.145
        0.145
        0.144
        0.145
        0.146
        0.147
        0.148
        0.148
        0.149
        0.15
        0.151
        0.15
        0.15
        0.152
        0.152
        0.153
        0.152
        0.152
        0.153
        0.154
        0.154
        0.154
        0.155
        0.154
        0.156
        0.155
        0.155
        0.155
        0.156
        0.156
        0.156
        0.157
        0.157
        0.157
        0.158
        0.157
        0.158
        0.157
        0.158
        0.158
        0.158
        0.16
        0.159
        0.159
        0.159
        0.159
        0.159
        0.159
        0.159
        0.159
        0.161
        0.16
        0.16
        0.159
        0.16
        0.16
        0.16
        0.16
        0.16
        0.16
        0.161
        0.16
        0.161
        0.161
        0.161
        0.161
        0.161
        0.161
        0.161
        };
        \addlegendentry{Small (8M) BiLSTM Hypernetwork}
      \end{axis}
    \end{tikzpicture}
    \caption{Comparison of large (210M parameters) and small (8M parameters) BiLSTM-based hypernetworks (HN-X-400) in terms of validation cosine similarity between predicted and true embeddings over 100 epochs for creating the \frameworkname-multi-400 model.}
    \label{fig:large_vs_small_hypernetwork}
\end{figure}
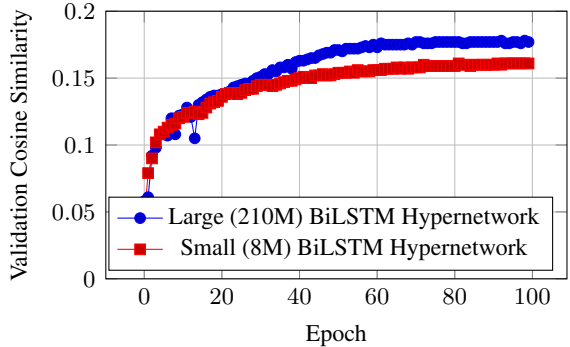

As described in \secref{sec:hypernetwork}, we use a BiLSTM architecture for hypernetworks. The hyperparameters of training are explained in the \secref{sec:hypernetwork_hyperparameters}.
Table~\ref{tab:hypernetwork_details} shows the hypernetwork parameter sizes used for each \frameworkname-based model.
Notably, the hypernetworks have a substantial number of parameters compared to their corresponding models. 
Preliminary experiments show that larger hypernetworks, when combined with strong regularization (dropout and the data augmentation methods), perform better than smaller hypernetworks.
Figure~\ref{fig:large_vs_small_hypernetwork} shows a case comparison study, which compares two hypernetworks for \frameworkname-multi-400 model, one with 210M and one with 8M parameters.
During training of the two hypernetworks, the larger one predicts embeddings better than the smaller one, when measuring cosine similarities to the true token embeddings in the validation set.
Also, as the dimension of the predicted embedding increases, a hypernetwork with higher capacity is necessary. 
Therefore, the hidden dimension of the BiLSTM is increased for embeddings with higher dimensions (see Appendix Table \ref{tab:hypernetwork_more_details}).

\begin{table}
    \small
    \setlength{\tabcolsep}{2pt}
    \centering
    \begin{tabular}{cc|cc}
        \toprule
        \textbf{LM} & 
        \textbf{Param} &
        \textbf{Hypernetwork} & \textbf{Param} \\
        \midrule
        \frameworkname-mono-100 & 92M & HN-R-100 & 22M \\
        \frameworkname-mono-200 & 97M & HN-R-200 & 23M \\
        \frameworkname-mono-400 & 107M & HN-R-400 & 87M \\
        \midrule
        \frameworkname-multi-100 & 113M & HN-X-100 & 53M \\
        \frameworkname-multi-100 & 138M & HN-X-200 & 54M \\
        \frameworkname-multi-400 & 188M & HN-X-400 & 210M \\
        \bottomrule
    \end{tabular}
    \caption{Number of parameters in \frameworkname-based models and their associated hypernetworks.}
 % Originally, RoBERTa has 125M and XLM-R has 280M parameters, but after factorization their embedding size is decreased.}
    \label{tab:hypernetwork_details}
\end{table}

\subsection{Baselines}

We consider the following baselines for comparison with \frameworkname. 
The details of how many tokens are randomly initialized or wisely initialized in each model are shown in Table~\ref{tab:init_token_volumes}.

\paragraph{\ofa-mono-xxx}
RoBERTa models \citep{liu2019roberta} with an extended vocabulary (from the original 50K to 401K) where the new token embeddings are initialized with \ofa \citep{liu-etal-2024-ofa}.

\paragraph{\ofa-multi-xxx}
XLM-R models \citep{conneau-etal-2020-unsupervised} with an extended vocabulary (from the original 250K to 401K) where the new token embeddings are initialized with \ofa \citep{liu-etal-2024-ofa}.

\paragraph{Random-mono-xxx}
RoBERTa models \citep{liu2019roberta} with an extended vocabulary (from the original 50K to 401K). Embeddings of all overlapping tokens are directly copied, while embeddings of the remaining tokens are randomly initialized from a Gaussian distribution with mean and standard deviations of the source embeddings.

\paragraph{Random-multi-xxx}
XLM-R models \citep{conneau-etal-2020-unsupervised} with an extended vocabulary (from the original 50K to 401K). Embeddings of all overlapping tokens are directly copied, while embeddings of the remaining tokens are randomly initialized from a Gaussian distribution with mean and standard deviations of the source embeddings.

\begin{table}[h!]
    \centering
    \small
    \begin{tabular}{lcccc}
        \toprule
        \textbf{Method} & \textbf{Model} & \textbf{Wise} & \textbf{Random} & \textbf{Total} \\
        \midrule
        \multirow{2}{*}{\frameworkname} & RoBERTa & 179K & 195K & 401K \\
        & XLM-R & 84K & 62K & 401K \\
        \cline{1-5}
        \multirow{2}{*}{\ofa} & RoBERTa & 179K & 195K & 401K \\
        & XLM-R & 84K & 62K & 401K \\
        \cline{1-5}
        \multirow{2}{*}{Random} & RoBERTa & 0 & 374K & 401K \\
        & XLM-R & 0 & 146K & 401K \\
        \bottomrule
    \end{tabular}
    \caption{Distribution of token embeddings initialized using \frameworkname, \ofa, and random initialization methods. 
    The ``Wise'' column indicates the number of tokens initialized using the respective wise initialization method. 
    The ``Random'' column indicates tokens initialized randomly. 
    The difference between the total tokens (``Total'') and the sum of ``Wise'' and ``Random'' columns represents token embeddings directly copied from the source embedding matrix due to vocabulary overlapping.
    This distribution holds consistently across all variants with different embedding factorization dimensions (100, 200, 400). Many token embeddings in \frameworkname and \ofa are wisely initialized.}
    \label{tab:init_token_volumes}
\end{table}

\subsection{Downstream Tasks}

The performances of \frameworkname-based models and the baselines are evaluated by four datasets in two downstream tasks: sentence retrieval and two sequence labeling, introduced as follows.

\paragraph{Sentence Retrieval}

Retrieval performance is assessed using the Sentence Retrieval Tatoeba (SR-T) \citep{artetxe-schwenk-2019-massively} and Sentence Retrieval Bible (SR-B) datasets. 
Following \citet{liu-etal-2024-ofa}, Top-10 accuracy is used as the evaluation metric, 
where the correct translation must be among the ten nearest neighbors of a query English sentence. 
Sentence-level representations are obtained by averaging contextualized word embeddings from the model's 8th layer.

\paragraph{Sequence Labeling}
For sequence labeling, named entity recognition (NER) and part-of-speech tagging (POS) are evaluated using WikiANN \citep{pan-etal-2017-cross} and Universal Dependencies \citep{de-marneffe-etal-2021-universal} datasets, respectively. 
Our evaluation methodology follows \citet{liu-etal-2024-ofa}, where models are fine-tuned on the English training set. 
The best checkpoint, selected based on validation performance, is then used to report zero-shot cross-lingual transfer performance on test sets in other languages. 
F1 scores are reported for both datasets.

\section{Results}

To validate the effectiveness of \frameworkname, we evaluate \frameworkname-based models and baselines in two scenarios: \textbf{before} (cf. \secref{sec:perf_before}) and \textbf{after} (cf. \secref{sec:perf_after}) the continual pre-training.

% The embedding matrices of LMs are extended, and the newly added tokens are initialized using three different methods: Random, \ofa, and \frameworkname. The performance of those LMs are compared under two conditions: before and after continued pre-training.

\subsection{Before Continual Pre-Training}\seclabel{sec:perf_before}

This evaluation aims to directly reflect the quality of the embeddings initialized with \frameworkname. Since the newly added tokens cover more than 500 languages (we use the Glot500-m tokenizer as the target tokenizer), we evaluate \frameworkname-based models and baselines on \textbf{all} languages in downstream tasks. The results are presented in Table~\ref{tab:performance_wo_continued_pre-training}.

\setlength{\tabcolsep}{4pt} % Reduce column spacing
\begin{table}[h!]
    \centering
    \begin{tabular}{lcccc}
        \toprule
        \textbf{Models} & \textbf{SR-B} & \textbf{SR-T} & \textbf{NER} & \textbf{POS} \\
        \midrule
        {\small Random-mono-100} & 3.5 & 4.6 & 23.4 & 22.5 \\
        {\small \ofa-mono-100} & 4.5 & 6.2 & \textbf{25.0} & \textbf{23.5} \\
        {\small \frameworkname-mono-100} & \textbf{5.0} & \textbf{6.4} & 24.9 & 22.8 \\
        \cline{1-5}
        {\small Random-mono-200} & 3.7 & 5.2 & 24.9 & 23.1 \\
        {\small \ofa-mono-200} & 4.5 & 7.2 & \textbf{25.7} & \textbf{23.4} \\
        {\small \frameworkname-mono-200} & \textbf{4.8} & \textbf{7.5} & 25.3 & \textbf{23.4} \\
        \cline{1-5}
        {\small Random-mono-400} & 4.1 & 5.3 & 25.8 & 23.0 \\
        {\small \ofa-mono-400} & \textbf{4.8} & \textbf{7.2} & \textbf{26.1} & \textbf{24.5} \\
        {\small \frameworkname-mono-400} & 4.7 & 6.3 & 25.8 & 23.0 \\
        \midrule
        \midrule
        {\small Random-multi-100} & 5.1 & 7.2 & 34.7 & 41.5 \\
        {\small \ofa-multi-100} & 5.1 & 7.5 & 36.3 & \textbf{42.3} \\
        {\small \frameworkname-multi-100} & \textbf{5.2} & \textbf{7.6} & \textbf{37.6} & \textbf{42.3} \\
        \cline{1-5}
        {\small Random-multi-200} & 5.7 & 10.0 & 38.1 & 47.3 \\
        {\small \ofa-multi-200} & 5.7 & 10.4 & \textbf{40.2} & \textbf{48.6} \\
        {\small \frameworkname-multi-200} & \textbf{6.0} & 1\textbf{0.6} & 38.3 & 48.3 \\
        \cline{1-5}
        {\small Random-multi-400} & 5.6 & 21.0 & 41.6 & 53.7 \\
        {\small \ofa-multi-400} & 5.9 & \textbf{21.3} & 43.3 & \textbf{54.6} \\
        {\small \frameworkname-multi-400} & \textbf{6.1} & \textbf{21.3} & \textbf{43.5} & 54.1 \\
        \bottomrule
    \end{tabular}
    \caption{Performance of randomly initialized baselines, \ofa and \frameworkname before continual pre-training. The scores for \ofa models are taken from \citet{liu-etal-2024-ofa} directly. SR-B covers \textbf{98} languages, SR-T covers \textbf{369} languages, NER covers \textbf{164} languages, and POS covers \textbf{91} languages. Top-10 accuracy is reported for SR-B and SR-T; F1 score is reported for NER and POS. All metrics are average across languages. 
    % \frameworkname achieves better performance than random baselines and its results are competitive with \ofa.
    }
    \label{tab:performance_wo_continued_pre-training}
\end{table}

\paragraph{\frameworkname and \ofa consistently outperform the random baselines, while showing comparable performance to each other across downstream tasks.}
In all downstream tasks,  the models with randomly initialized new embeddings perform the worst.
This indicates that randomly initializing the new token embeddings is suboptimal as no encoded knowledge in the original embedding matrix is explicitly leveraged.
For the retrieval tasks (SR-B and SR-T), \frameworkname performs better than \ofa on all cases except when the embedding dimension is 400 in the mono setup. 
We hypothesize this might be because, with a fixed amount of training data (22K pairs for mono models), learning higher-dimensional embeddings becomes more challenging for the hypernetwork.
This hypothesis is supported by the fact that when more training instances are included in the multi models (103 pairs), \frameworkname-mutli-400 models achieve comparable or even better results than \ofa-multi-400 models across all downstream tasks.
% because increasing the embedding dimension (from 100 to 200 and 400) leads to a decline in \frameworkname’s performance.
% With a fixed amount of training data, learning higher-dimensional embeddings becomes more challenging for the hypernetwork. 
% Nevetherless, 
% In mono models, where the hypernetwork had only 20K training observations, \frameworkname-mono-400 underperforms compared to OFA-mono-400 across all four benchmarks. 
% However, \frameworkname-multi-400 performs competitively with OFA-multi-400, as the hypernetwork learned with 100K training observations.
% For sequence labeling tasks (NER and POS), \ofa generally outperforms \frameworkname, though \frameworkname occasionally achieves better results.

\subsection{After Continual Pre-Training}\seclabel{sec:perf_after}

Continual pre-training is crucial because, even with carefully initialized new token embeddings, the embeddings and the backbone model must be fine-tuned on data containing these new tokens. 
% Additionally, the model parameters need to adapt to these languages.
Therefore, to validate how effective the new embeddings with \frameworkname are as a starting point for continual pre-training, we select 6 models and continually pre-train them on a diverse set of languages.

\paragraph{Models and Training}
Due to resource constraints, we select \textbf{6} models out of 18 models for continual pre-training. 
For the mono models, we use Random-mono-100, \ofa-mono-100, and \frameworkname-mono-100; for the multi models, we use Random-multi-400, \ofa-multi-400, and \frameworkname-multi-400.
All six models are continually pre-trained using hyperparameters similar to those in \citet{liu-etal-2024-ofa}, with some key differences: an effective batch size of 512 instead of 384 and training on 4 NVIDIA H100 GPUs. 
The training is conducted for 4,000 steps (approx. 1 epoch).
% The results of training loss and retrieval benchmark evaluations are presented in Figure \ref{fig:continued_pre-trainig_curves}.

\paragraph{Training Data} Due to constrained computing resources, we are not able to continually train \frameworkname-based models or other baselines on full Glot500-c \citep{imanigooghari-etal-2023-glot500}. Therefore, a subset of languages from Glot500-c comprising \textbf{22} languages spanning high, mid, and low-resource categories is used for the continual pre-training.
% For continued pre-training, a subset of languages from Glot500-c \citep{imanigooghari-etal-2023-glot500}, comprising 22 languages spanning high, mid, and low-resource categories is used. 
The list of languages and their data size can be found in Appendix Table~\ref{tab:language-data}. This dataset subset contains 1.1 billion tokens across 36 million sentences.

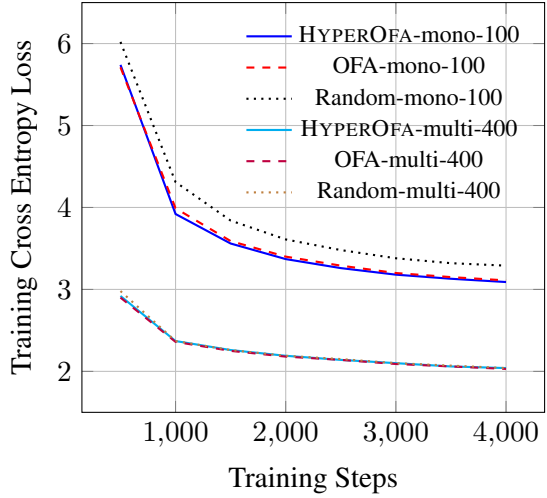
\begin{figure}[h!]
    \centering
    \begin{tikzpicture}
        \begin{axis}[
            grid=major,
            ymin=1.5, ymax=6.5,
            xmode=linear,
            ymode=linear,
            width=1.0\linewidth,  % wider
            height=7cm,            % taller
            xlabel={Training Steps},
            ylabel={Training Cross Entropy Loss},            
            legend style={
                font=\small,
                at={(0.97,0.97)},
                anchor=north east,
                draw=none,
                fill=none,
                inner sep=2pt
            }
        ]
        \addplot[solid, blue, thick] coordinates {
            (500, 5.74) (1000, 3.92) (1500, 3.56) (2000, 3.37)
            (2500, 3.26) (3000, 3.18) (3500, 3.13) (4000, 3.09)
        };
        \addlegendentry{\frameworkname-mono-100}
        
        \addplot[dashed, red, thick] coordinates {
            (500, 5.71) (1000, 3.99) (1500, 3.59) (2000, 3.40)
            (2500, 3.29) (3000, 3.20) (3500, 3.15) (4000, 3.11)
        };
        \addlegendentry{OFA-mono-100}
        
        \addplot[dotted, black, thick] coordinates {
            (500, 6.02) (1000, 4.31) (1500, 3.84) (2000, 3.61)
            (2500, 3.48) (3000, 3.38) (3500, 3.32) (4000, 3.29)
        };
        \addlegendentry{Random-mono-100}
        
        \addplot[solid, cyan, thick] coordinates {
            (500, 2.92) (1000, 2.37) (1500, 2.26) (2000, 2.19)
            (2500, 2.14) (3000, 2.10) (3500, 2.06) (4000, 2.04)
        };
        \addlegendentry{\frameworkname-multi-400}
        
        \addplot[dashed, purple, thick] coordinates {
            (500, 2.90) (1000, 2.36) (1500, 2.25) (2000, 2.18)
            (2500, 2.14) (3000, 2.09) (3500, 2.06) (4000, 2.03)
        };
        \addlegendentry{OFA-multi-400}
        
        \addplot[dotted, brown, thick] coordinates {
            (500, 2.98) (1000, 2.37) (1500, 2.26) (2000, 2.19)
            (2500, 2.15) (3000, 2.10) (3500, 2.07) (4000, 2.04)
        };
        \addlegendentry{Random-multi-400}
        
        \end{axis}
    \end{tikzpicture}
    \caption{Training loss curves during the continual pre-training of models initialized with \frameworkname, \ofa, or random initialization methods.}
    \label{fig:training_loss_curves}
\end{figure}

The benchmark results for before and after continual pre-training for the 6 models are presented in Table \ref{tab:before-after-cont-training-results}. The metrics are calculated for the languages that are in the 22 continual pre-training languages. And the training loss curves of the 6 models throughout the continual pre-training are presented in Figure \ref{fig:training_loss_curves}.

\begin{table}
  \centering
  \small
  \begin{tabular}{l l@{\hskip 6pt}r@{\hskip 3pt}r@{\hskip 3pt}r@{\hskip 3pt}r}
    \toprule
    \textbf{Model}       & \textbf{Phase}
      & \textbf{SRT} & \textbf{SRB}
      & \textbf{POS} & \textbf{NER} \\
    \midrule
    \multirow{2}{*}{Random-mono-100}
                         & Before &  4.4 &  3.6 & 29.1 & 23.3 \\
                         & After  &  9.5 &  7.0 & 51.1 & 40.0 \\ 

    \addlinespace
    \multirow{2}{*}{\ofa-mono-100}
                         & Before &  5.9 &  5.0 & 30.2 & 24.0 \\
                         & After  & \textbf{15.2} &  9.8 & \textbf{56.8} & \textbf{45.7} \\      

    \addlinespace                         
    \multirow{2}{*}{\frameworkname-mono-100}
                         & Before &  6.0 &  5.1 & 30.0 & 23.5 \\
                         & After  & 11.3 &  \textbf{9.9} & 56.3 & 43.4 \\
    \midrule
    \multirow{2}{*}{Random-multi-400}
                         & Before & 17.6 &  8.1 & 65.0 & 45.9 \\
                         & After  & 55.3 & 40.8 & 70.3 & 59.8 \\

    \addlinespace
    \multirow{2}{*}{\ofa-multi-400}
                         & Before & 17.9 &  8.6 & 62.9 & 47.2 \\
                         & After  & 55.8 & \textbf{42.3} & \textbf{70.4} & 60.3 \\                 
    
    \addlinespace
    \multirow{2}{*}{\frameworkname-multi-400}
                         & Before & 17.7 &  9.2 & 63.7 & 47.5 \\
                         & After  & \textbf{56.1} & 42.2 & \textbf{70.4} & \textbf{60.5} \\                             
    \bottomrule
  \end{tabular}
  \caption{Performance before and after continual pre-training. Evaluation is conducted on the intersection of the 22 continual pre-training languages and those available in each downstream task.
  Specifically, SR-T and SR-B are evaluated on \textbf{20} languages, POS on \textbf{9} languages, and NER on \textbf{14} languages. Metrics reported are: Top-10 accuracy for SR-T and SR-B, F1 score for POS NER. All metrics are averaged across the respective languages. \frameworkname achieves consistently better performance than the random baseline and competitive performance compared with \ofa.}
  \label{tab:before-after-cont-training-results}
\end{table}

\paragraph{Multilingual XLM-R models consistently outperform their monolingual RoBERTa counterparts, highlighting the advantages of multilingual pre-training.}
The first observation is that all models based on XLM-R outperform the RoBERTa-based models. 
This aligns with our expectations, as XLM-R already sees much multilingual data during its pre-training stage, which helps further adapt to other languages.
In contrast, RoBERTa is originally monolingual and therefore lacks enough multilingual knowledge.

\paragraph{Within XLM-R models, the choice of embedding initialization has minimal impact, suggesting inherent robustness to vocabulary extension.}
Different initialization (random, \ofa, or \frameworkname) methods do not produce substantial performance differences in models based on XLM-R across downstream tasks. The loss curves (cf. Figure \ref{fig:training_loss_curves}) also show that different multilingual models show a similar convergence trend throughout continual pre-training progression.
This suggests that multilingual models are already quite robust and effective in adapting to new languages even when new token embeddings are randomly initialized.

\paragraph{RoBERTa-based models benefit from wise initialization methods.}
Models with embeddings initialized using \ofa and \frameworkname show notably improved performance compared to those with the random baseline in RoBERTa-based models across all downstream tasks. Additionally, \ofa and \frameworkname also show faster convergence (at the same training step but a lower loss) than the random baseline, as shown in Figure \ref{fig:training_loss_curves}.
This highlights the significance of advanced embedding initialization techniques for monolingual models -- a better strategy can actively leverage the knowledge encoded in the original embeddings, though monolingual, and can be transferred to other languages.

\paragraph{\frameworkname and \ofa perform comparably across downstream tasks, suggesting both are viable strategies.}
We observe that \frameworkname achieves comparable or occasionally better results than \ofa.
However, the difference is generally small, with neither method showing a decisive advantage overall. 
This suggests that both approaches are effective, with their relative strengths depending on the specific evaluation metric.
However, because of the capability of modeling non-linearity, we expect \frameworkname-based models can improve when more training data (for hypernetworks and continual pre-training) is available.

% Overall, the advanced initialization techniques - \frameworkname and \ofa methods - employed to extend the vocabulary of mono-lingual RoBERTa demonstrate superior performance compared to random initialization in scenarios both with and without continued pre-training. This reduces computational costs and time requirements while achieving strong performance when extending language models to accommodate new languages. In contrast, for the multilingual XLM-R model, while these wise initialization techniques outperform random initialization in setups without continued pre-training, they yield no substantial advantage in scenarios involving continued pre-training.

\section{Conclusion}

This study introduces \frameworkname, a method for expanding the vocabulary of PLMs to new languages and initializing new token embeddings with a hypernetwork.
% \frameworkname leverages external static multilingual word vectors as an external information source to initialize new token embeddings.
We show the effectiveness of \frameworkname by evaluating the resulting models both before and after the continual pre-training.
The results show that \frameworkname consistently outperforms the random initialization baseline and performs competitively with \ofa. 
These results highlight \frameworkname as a promising approach, alongside \ofa, for efficient new token embedding initialization towards effective and efficient continual pre-training.

\section*{Limitations}

This study explores initializing new embeddings in encoder-only models. 
While both methods are theoretically applicable to decoder-only models like GPT \citep{radford2019language} and encoder-decoder models like T5 \citep{raffel2020exploring}, the effectiveness in these settings remains untested, presenting an open research direction.

Another limitation concerns the embedding dimensions used in this study. Due to the embedding matrix factorization described in \secref{sec:factorization}, the dimensions are relatively low compared to those in modern LLMs. While this approach reduces computational costs, it leaves open the question of how \frameworkname would perform with much higher-dimensional embeddings.

Finally, the continual pre-trained dataset used in this study is relatively small compared to that of \citet{liu-etal-2024-ofa} due to computational constraints. 
Exploring the impact of larger datasets, especially those having more languages, could provide deeper insights into the strengths and weaknesses of the proposed methods in different settings.

\section*{Acknowledgements}
We sincerely thank Mina Rezaei for insightful discussions. We also gratefully acknowledge the Leibniz Supercomputing Centre (LRZ) of the Bavarian Academy of Sciences and Humanities and the Munich Center for Machine Learning (MCML) for generously providing computational resources.

\bibliography{custom}

\appendix

\section{Experiments for Hypernetwork}

\subsection{Architecture: BiLSTM vs Setformer}
\seclabel{sec:hypernetwork_architecture}

As explained in the \secref{sec:hypernetwork}, there are two requirements for the model architecture; variable length input, permutation invariant. To satisfy those requirements, initially, an encoder only transformer model \citep{vaswani2017attention} without positional encoding layers (called as Setformer in this study) was tested. However, after observing poor performance, the approach shifted to a BiLSTM (Bidirectional LSTM) architecture \citep{schuster1997bidirectional} despite it not inherently satisfying the permutation-invariance requirement. Experimental results demonstrated that BiLSTM works better for this task when compared to a transfomer encoder model without positional encoding layer (Table \ref{tab:setformer_vs_bilstm}).

Table~\ref{tab:setformer_vs_bilstm} compares the two candidate hypernetwork architectures, Setformer and BiLSTM, for initializing token embeddings for \frameworkname-mono-100 model. The model initialized with the BiLSTM hypernetwork achieves better SR-T Top 10 accuracy (6.4), outperforming the Setformer variant. This suggests that BiLSTM is more effective than Setformer as a hypernetwork. 

We attribute the reason for the poor performance of the Setformer to the need of transformers that require a large amount of data to learn effectively. On the other hand, the BiLSTM architecture was more efficient at learning the task with the available data which is limited by the source vocabulary size.

\begin{table}[h!]
    \centering
    % \small
    \begin{tabular}{cccc}
        \toprule
        \textbf{LM} & \textbf{Hypernetwork} & \textbf{SR-T} \\
        \midrule
        % \small{\frameworkname-mono-100} & \small{Setformer} & 8M & 5.7 \\
        % \small{\frameworkname-mono-100} & \small{BiLSTM} & 7M & 5.6 \\
        % \midrule
        \frameworkname-mono-100 & BiLSTM & \textbf{6.4} \\
        \frameworkname-mono-100 & Setformer & 5.2 \\
        \midrule
        Random-mono-100 & - & 4.6 \\
        \bottomrule
    \end{tabular}
    \caption{Comparison of Setformer (Transformer encoder without positional encodings) and BiLSTM as hypernetworks both having 22M trainable parameters. They are used for initializing token embeddings in \frameworkname-mono-100, a RoBERTa-based model with a new vocabulary and factorized embedding dimension of 100 (mono-100). The SR-T Top 10 Accuracy is reported for the without continual pre-training set up. Random initialization baseline performance is given at the last row. BiLSTM performs better as a hypernetwork.}
    \label{tab:setformer_vs_bilstm}
\end{table}

\subsection{Hyperparameters}\seclabel{sec:hypernetwork_hyperparameters}

The hypernetworks follow a BiLSTM architecture. All hypernetworks for \frameworkname-mono-xxx and \frameworkname-multi-xxx models share the same configuration: a maximum context size of 256, a dropout rate of 0.4, and an Adam optimizer. The learning rate starts at \(1 \times 10^{-4}\) and decays linearly by a factor of 0.95 every 10 epochs. Training was conducted on two Nvidia A100 GPUs, with each model requiring approximately 1 to 1.5 hours. 

To ensure a healthy training, the hyperparameters in the loss function, as explained in \secref{sec:hypernetwork}, were set as follows: $\lambda = 0.1$ for all hypernetworks, and $T = 0.5$ for the hypernetworks of \frameworkname-mono-xxx, and $T = 0.25$ for the hypernetworks of \frameworkname-multi-xxx.

All models were trained until the validation loss converged. More details about the training data, model parameter sizes are presented in Table \ref{tab:hypernetwork_more_details}.

\begin{table*}[h!]
    \centering
    \begin{tabular}{lcccccc}
        \toprule
        \textbf{LM} & \textbf{Hypernetwork} & \textbf{Training Data} & \textbf{Layers} & \textbf{Hid Dim} & \textbf{Param} & \textbf{Epoch} \\
        \midrule
        \frameworkname-mono-100 & HN-R-100 & 22K & 2 & 800 & 22M & 370 \\
        \frameworkname-mono-200 & HN-R-200 & 22K & 2 & 800 & 23M & 470 \\
        \frameworkname-mono-400 & HN-R-400 & 22K & 2 & 1600 & 87M & 400 \\
        \midrule
        \frameworkname-multi-100 & HN-X-100 & 103K & 4 & 800 & 53M & 120 \\
        \frameworkname-multi-200 & HN-X-200 & 103K & 4 & 800 & 54M & 230 \\
        \frameworkname-multi-400 & HN-X-400 & 103K & 4 & 1600 & 210M & 80 \\
        \bottomrule
    \end{tabular}
    \caption{Hypernetwork model details for predicting the target embeddings for \frameworkname-mono-xxx and \frameworkname-multi-xxx language models with different factorized dimensions. All hypernetworks have the BiLSTM architecture. Epochs column indicated the converged epoch number for the hypernetwork.}
    \label{tab:hypernetwork_more_details}
\end{table*}

\subsection{Regularization}

We applied multiple regularization and data augmentation methods to ensure that hypernetworks do not overfit. 

We used high dropout rate of 0.4 since we have seen that the large models with high regularization performs better (see Figure \ref{fig:large_vs_small_hypernetwork}). We also applied data augmentation by shuffling word vector order before each epoch to prevent model to overfit to the order of the input word vectors.

Additionally, with 50\% probability, the number of word vectors is randomly limited to 50–100\% of the available vectors.

\section{Continual Pre-training Dataset}

The continual pre-training dataset was deliberately kept smaller than that used by \citet{liu-etal-2024-ofa} due to disk quota limitations in the \frameworkname study. The languages, their original sentence counts in Glot500-c \citep{imanigooghari-etal-2023-glot500} dataset and the sentence counts used in this study is listed in Table~\ref{tab:language-data}. For continual pre-training 36M sentences (approx. 1.1B tokens) across 22 languages are used. To categorize source category with respect to the volume of that language in Glot500-c, thresholds used: high (>5M sentences), mid (>500K sentences), and low (<500K sentences).

\begin{table*}[h!]
\centering
\small
\begin{tabular}{llrr}
\toprule
\textbf{Source Category} & \textbf{Language} & \textbf{Glot500-c Sentence Count} & \textbf{Subsampled Sentence Count} \\
\midrule
\multirow{6}{*}{High} 
& eng\_Latn & 36,121,560 & 5,000,000 \\
& tur\_Latn & 29,182,577 & 5,000,000 \\
& ell\_Grek & 22,031,905 & 5,000,000 \\
& bul\_Cyrl & 21,822,051 & 5,000,000 \\
& ces\_Latn & 20,374,860 & 5,000,000 \\
& kor\_Hang & 6,348,091 & 5,000,000 \\
\cmidrule(l){2-4}
\multirow{5}{*}{Mid}
& kat\_Geor & 990,785 & 990,785 \\
& fry\_Latn & 925,801 & 925,801 \\
& zsm\_Latn & 849,033 & 849,033 \\
& khm\_Khmr & 565,794 & 565,794 \\
& jpn\_Japn & 507,538 & 507,538 \\
\cmidrule(l){2-4}
\multirow{11}{*}{Low}
& yue\_Hani & 483,750 & 483,750 \\
& tuk\_Latn & 312,480 & 312,480 \\
& uig\_Arab & 298,694 & 298,694 \\
& pam\_Latn & 292,293 & 292,293 \\
& kab\_Latn & 166,953 & 166,953 \\
& gla\_Latn & 124,953 & 124,953 \\
& mhr\_Cyrl & 91,557 & 91,557 \\
& swh\_Latn & 43,876 & 43,876 \\
& cmn\_Hani & 57,500 & 57,500 \\
& pes\_Arab & 18,762 & 18,762 \\
& dtp\_Latn & 1,355 & 1,355 \\
\midrule[\heavyrulewidth]
\multicolumn{2}{r}{\textit{Total Sentence Count}} & 141,612,168 & 35,731,124
\end{tabular}
\caption{Distribution of continued pre-trainig data. The table shows the original Glot500-c volume and sub-sampled volume for each language, grouped by their source category (High, Mid, Low) which is assigned with respect to the volume of that language in Glot500-c.}
\label{tab:language-data}
\end{table*}

\section{Benchmark Language Coverage}

In this section, we present the languages used in benchmarks for the tables in our paper.

\subsection{For Benchmark Performances in Table~\ref{tab:performance_wo_continued_pre-training}}

SR-B Benchmark Languages:

{\small mal\_Mlym, aze\_Latn, guj\_Gujr, ben\_Beng, kan\_Knda, tel\_Telu, mlt\_Latn, fra\_Latn, spa\_Latn, fil\_Latn, nob\_Latn, rus\_Cyrl, deu\_Latn, tur\_Latn, pan\_Guru, mar\_Deva, por\_Latn, nld\_Latn, zho\_Hani, ita\_Latn, ind\_Latn, ell\_Grek, bul\_Cyrl, swe\_Latn, ces\_Latn, isl\_Latn, pol\_Latn, ron\_Latn, dan\_Latn, hun\_Latn, tgk\_Cyrl, srp\_Latn, fas\_Arab, ceb\_Latn, heb\_Hebr, hrv\_Latn, fin\_Latn, slv\_Latn, vie\_Latn, mkd\_Cyrl, slk\_Latn, nor\_Latn, est\_Latn, ltz\_Latn, eus\_Latn, lit\_Latn, kaz\_Cyrl, lav\_Latn, epo\_Latn, cat\_Latn, tha\_Thai, ukr\_Cyrl, tgl\_Latn, sin\_Sinh, gle\_Latn, hin\_Deva, kor\_Hang, ory\_Orya, urd\_Arab, sqi\_Latn, bel\_Cyrl, afr\_Latn, nno\_Latn, tat\_Cyrl, hau\_Latn, sna\_Latn, msa\_Latn, som\_Latn, srp\_Cyrl, mlg\_Latn, zul\_Latn, arz\_Arab, nya\_Latn, tam\_Taml, hat\_Latn, uzb\_Latn, sot\_Latn, uzb\_Cyrl, als\_Latn, amh\_Ethi, sun\_Latn, war\_Latn, yor\_Latn, fao\_Latn, uzn\_Cyrl, smo\_Latn, bak\_Cyrl, ilo\_Latn, tso\_Latn, mri\_Latn, asm\_Beng, hil\_Latn, nso\_Latn, ibo\_Latn, kin\_Latn, hye\_Armn, lin\_Latn, tpi\_Latn, twi\_Latn, kir\_Cyrl, pap\_Latn, nep\_Deva, bcl\_Latn, xho\_Latn, cym\_Latn, gaa\_Latn, ton\_Latn, lat\_Latn, srn\_Latn, ewe\_Latn, bem\_Latn, efi\_Latn, bis\_Latn, haw\_Latn, hmo\_Latn, kat\_Geor, pag\_Latn, loz\_Latn, fry\_Latn, mya\_Mymr, nds\_Latn, run\_Latn, rar\_Latn, fij\_Latn, ckb\_Arab, ven\_Latn, zsm\_Latn, chv\_Cyrl, sag\_Latn, guw\_Latn, bre\_Latn, toi\_Latn, che\_Cyrl, pis\_Latn, oss\_Cyrl, nan\_Latn, tuk\_Latn, tir\_Ethi, yua\_Latn, min\_Latn, khm\_Khmr, tum\_Latn, lug\_Latn, tzo\_Latn, mah\_Latn, jav\_Latn, jpn\_Jpan, lus\_Latn, crs\_Latn, ndo\_Latn, snd\_Arab, yue\_Hani, kua\_Latn, hin\_Latn, kal\_Latn, tdt\_Latn, mfe\_Latn, mos\_Latn, kik\_Latn, cnh\_Latn, gil\_Latn, pon\_Latn, ori\_Orya, luo\_Latn, nzi\_Latn, gug\_Latn, bar\_Latn, bci\_Latn, chk\_Latn, yap\_Latn, ssw\_Latn, quz\_Latn, sah\_Cyrl, tsn\_Latn, quy\_Latn, bbc\_Latn, wal\_Latn, uig\_Arab, pam\_Latn, seh\_Latn, zai\_Latn, gym\_Latn, bod\_Tibt, nde\_Latn, fon\_Latn, nbl\_Latn, kmr\_Latn, guc\_Latn, mam\_Latn, nia\_Latn, nyn\_Latn, cab\_Latn, top\_Latn, mco\_Latn, tzh\_Latn, plt\_Latn, iba\_Latn, kek\_Latn, sop\_Latn, kac\_Latn, qvi\_Latn, cak\_Latn, kbp\_Latn, ctu\_Latn, kri\_Latn, mau\_Latn, tyv\_Cyrl, btx\_Latn, nch\_Latn, ncj\_Latn, pau\_Latn, toj\_Latn, pcm\_Latn, dyu\_Latn, kss\_Latn, quc\_Latn, yao\_Latn, kab\_Latn, tuk\_Cyrl, ndc\_Latn, san\_Deva, qug\_Latn, arb\_Arab, mck\_Latn, arn\_Latn, pdt\_Latn, gla\_Latn, kmr\_Cyrl, nav\_Latn, ksw\_Mymr, mxv\_Latn, hif\_Latn, wol\_Latn, sme\_Latn, gom\_Latn, bum\_Latn, mgr\_Latn, ahk\_Latn, tsz\_Latn, bzj\_Latn, udm\_Cyrl, cce\_Latn, meu\_Latn, cbk\_Latn, bhw\_Latn, ngu\_Latn, nyy\_Latn, naq\_Latn, toh\_Latn, nse\_Latn, alz\_Latn, mhr\_Cyrl, djk\_Latn, gkn\_Latn, grc\_Grek, swh\_Latn, alt\_Cyrl, miq\_Latn, kaa\_Cyrl, lhu\_Latn, lzh\_Hani, cmn\_Hani, kjh\_Cyrl, mgh\_Latn, rmy\_Latn, srm\_Latn, gur\_Latn, yom\_Latn, cfm\_Latn, lao\_Laoo, qub\_Latn, ote\_Latn, ldi\_Latn, ayr\_Latn, bba\_Latn, aln\_Latn, leh\_Latn, ban\_Latn, ace\_Latn, pes\_Arab, ary\_Arab, hus\_Latn, glv\_Latn, mai\_Deva, dzo\_Tibt, ctd\_Latn, nnb\_Latn, sxn\_Latn, mps\_Latn, gkp\_Latn, acr\_Latn, dtp\_Latn, lam\_Latn, poh\_Latn, quh\_Latn, tob\_Latn, ach\_Latn, npi\_Deva, myv\_Cyrl, tih\_Latn, gor\_Latn, ium\_Latn, teo\_Latn, kia\_Latn, crh\_Cyrl, enm\_Latn, mad\_Latn, cac\_Latn, hnj\_Latn, ikk\_Latn, sba\_Latn, zom\_Latn, bqc\_Latn, bim\_Latn, mdy\_Ethi, bts\_Latn, gya\_Latn, agw\_Latn, knv\_Latn, giz\_Latn, hui\_Latn, hif\_Deva}

SR-T Benchmark Languages:

{\small
mal\_Mlym, aze\_Latn, ben\_Beng, tel\_Telu, fra\_Latn, spa\_Latn, nob\_Latn, rus\_Cyrl, deu\_Latn, tur\_Latn, mar\_Deva, por\_Latn, nld\_Latn, ara\_Arab, ita\_Latn, ind\_Latn, ell\_Grek, bul\_Cyrl, swe\_Latn, ces\_Latn, isl\_Latn, pol\_Latn, ron\_Latn, dan\_Latn, hun\_Latn, srp\_Latn, ceb\_Latn, heb\_Hebr, hrv\_Latn, glg\_Latn, fin\_Latn, slv\_Latn, vie\_Latn, mkd\_Cyrl, slk\_Latn, est\_Latn, eus\_Latn, lit\_Latn, kaz\_Cyrl, bos\_Latn, epo\_Latn, cat\_Latn, tha\_Thai, ukr\_Cyrl, tgl\_Latn, gle\_Latn, hin\_Deva, kor\_Hang, urd\_Arab, sqi\_Latn, bel\_Cyrl, afr\_Latn, nno\_Latn, tat\_Cyrl, ast\_Latn, mon\_Cyrl, arz\_Arab, tam\_Taml, uzb\_Cyrl, amh\_Ethi, war\_Latn, fao\_Latn, hye\_Armn, oci\_Latn, xho\_Latn, cym\_Latn, lat\_Latn, kat\_Geor, fry\_Latn, nds\_Latn, zsm\_Latn, bre\_Latn, tuk\_Latn, khm\_Khmr, jpn\_Jpan, yue\_Hani, gsw\_Latn, lvs\_Latn, kur\_Latn, ido\_Latn, uig\_Arab, pam\_Latn, pms\_Latn, wuu\_Hani, yid\_Hebr, ina\_Latn, kab\_Latn, gla\_Latn, cbk\_Latn, hsb\_Latn, mhr\_Cyrl, swh\_Latn, cmn\_Hani, pes\_Arab, dtp\_Latn, lfn\_Latn, ile\_Latn, csb\_Latn.
}

NER Benchmark Languages:

{\small
hbs\_Latn, mal\_Mlym, aze\_Latn, guj\_Gujr, ben\_Beng, kan\_Knda, tel\_Telu, mlt\_Latn, fra\_Latn, spa\_Latn, eng\_Latn, rus\_Cyrl, deu\_Latn, tur\_Latn, pan\_Guru, mar\_Deva, por\_Latn, nld\_Latn, ara\_Arab, zho\_Hani, ita\_Latn, ind\_Latn, ell\_Grek, bul\_Cyrl, swe\_Latn, ces\_Latn, isl\_Latn, pol\_Latn, ron\_Latn, dan\_Latn, hun\_Latn, tgk\_Cyrl, fas\_Arab, ceb\_Latn, heb\_Hebr, hrv\_Latn, glg\_Latn, fin\_Latn, slv\_Latn, vie\_Latn, mkd\_Cyrl, slk\_Latn, nor\_Latn, est\_Latn, ltz\_Latn, eus\_Latn, lit\_Latn, kaz\_Cyrl, lav\_Latn, bos\_Latn, epo\_Latn, cat\_Latn, tha\_Thai, ukr\_Cyrl, tgl\_Latn, sin\_Sinh, gle\_Latn, hin\_Deva, kor\_Hang, urd\_Arab, swa\_Latn, sqi\_Latn, bel\_Cyrl, afr\_Latn, nno\_Latn, tat\_Cyrl, ast\_Latn, mon\_Cyrl, msa\_Latn, som\_Latn, srp\_Cyrl, mlg\_Latn, arz\_Arab, tam\_Taml, uzb\_Latn, cos\_Latn, als\_Latn, amh\_Ethi, sun\_Latn, war\_Latn, div\_Thaa, yor\_Latn, fao\_Latn, bak\_Cyrl, ilo\_Latn, mri\_Latn, asm\_Beng, ibo\_Latn, kin\_Latn, hye\_Armn, oci\_Latn, lin\_Latn, kir\_Cyrl, nep\_Deva, cym\_Latn, lat\_Latn, kat\_Geor, fry\_Latn, mya\_Mymr, nds\_Latn, pnb\_Arab, ckb\_Arab, chv\_Cyrl, que\_Latn, bre\_Latn, pus\_Arab, che\_Cyrl, oss\_Cyrl, nan\_Latn, lim\_Latn, tuk\_Latn, min\_Latn, khm\_Khmr, jav\_Latn, vec\_Latn, jpn\_Jpan, snd\_Arab, yue\_Hani, sco\_Latn, ori\_Orya, arg\_Latn, kur\_Latn, bar\_Latn, roh\_Latn, aym\_Latn, sah\_Cyrl, lmo\_Latn, ido\_Latn, vol\_Latn, uig\_Arab, bod\_Tibt, pms\_Latn, wuu\_Hani, yid\_Hebr, scn\_Latn, ina\_Latn, xmf\_Geor, san\_Deva, gla\_Latn, mwl\_Latn, diq\_Latn, cbk\_Latn, szl\_Latn, hsb\_Latn, vls\_Latn, mhr\_Cyrl, grn\_Latn, lzh\_Hani, mzn\_Arab, nap\_Latn, ace\_Latn, frr\_Latn, eml\_Latn, vep\_Latn, sgs\_Latn, lij\_Latn, crh\_Latn, ksh\_Latn, zea\_Latn, csb\_Latn, jbo\_Latn, bih\_Deva, ext\_Latn, fur\_Latn.
}

POS Benchmark Languages:

{\small
mal\_Mlym, ben\_Beng, tel\_Telu, mlt\_Latn, fra\_Latn, spa\_Latn, eng\_Latn, rus\_Cyrl, deu\_Latn, tur\_Latn, mar\_Deva, por\_Latn, nld\_Latn, ara\_Arab, zho\_Hani, ita\_Latn, ind\_Latn, ell\_Grek, bul\_Cyrl, swe\_Latn, ces\_Latn, isl\_Latn, pol\_Latn, ron\_Latn, dan\_Latn, hun\_Latn, srp\_Latn, fas\_Arab, ceb\_Latn, heb\_Hebr, hrv\_Latn, glg\_Latn, fin\_Latn, slv\_Latn, vie\_Latn, slk\_Latn, nor\_Latn, est\_Latn, eus\_Latn, lit\_Latn, kaz\_Cyrl, lav\_Latn, cat\_Latn, tha\_Thai, ukr\_Cyrl, tgl\_Latn, sin\_Sinh, gle\_Latn, hin\_Deva, kor\_Hang, urd\_Arab, sqi\_Latn, bel\_Cyrl, afr\_Latn, tat\_Cyrl, tam\_Taml, amh\_Ethi, yor\_Latn, fao\_Latn, hye\_Armn, cym\_Latn, lat\_Latn, nds\_Latn, bre\_Latn, hyw\_Armn, jav\_Latn, jpn\_Jpan, yue\_Hani, gsw\_Latn, sah\_Cyrl, uig\_Arab, kmr\_Latn, pcm\_Latn, quc\_Latn, san\_Deva, gla\_Latn, wol\_Latn, sme\_Latn, hsb\_Latn, grc\_Grek, hbo\_Hebr, grn\_Latn, lzh\_Hani, ajp\_Arab, nap\_Latn, aln\_Latn, glv\_Latn, lij\_Latn, myv\_Cyrl, bam\_Latn, xav\_Latn.
}

\subsection{For Benchmark Performances in Table ~\ref{tab:before-after-cont-training-results}}

SR-T Benchmark Languages:

{\small
tur\_Latn, ell\_Grek, bul\_Cyrl, ces\_Latn, kor\_Hang, zsm\_Latn, kat\_Geor, fry\_Latn, khm\_Khmr, yue\_Hani, tuk\_Latn, uig\_Arab, pam\_Latn, kab\_Latn, gla\_Latn, mhr\_Cyrl, swh\_Latn, cmn\_Hani, pes\_Arab, dtp\_Latn
}

SR-B Benchmark Languages:

{\small
tur\_Latn, ell\_Grek, bul\_Cyrl, ces\_Latn, kor\_Hang, zsm\_Latn, kat\_Geor, fry\_Latn, khm\_Khmr, yue\_Hani, tuk\_Latn, uig\_Arab, pam\_Latn, kab\_Latn, gla\_Latn, mhr\_Cyrl, swh\_Latn, cmn\_Hani, pes\_Arab, dtp\_Latn
}

NER Benchmark Languages:

{\small
eng\_Latn, tur\_Latn, ell\_Grek, bul\_Cyrl, ces\_Latn, kor\_Hang, kat\_Geor, fry\_Latn, khm\_Khmr, yue\_Hani, tuk\_Latn, uig\_Arab, gla\_Latn, mhr\_Cyrl
}

POS Benchmark Languages:

{\small
eng\_Latn, tur\_Latn, ell\_Grek, bul\_Cyrl, ces\_Latn, kor\_Hang, yue\_Hani, uig\_Arab, gla\_Latn
}

\section{Performance - Language Breakdown}

In this section we show the benchmark results per language before continual pre-training (checkpoint 0) and after (checkpoint 4000) for the 6 models which had continual pre-training (see \secref{sec:perf_after}).

\begin{table*}[h]
    \centering
    \footnotesize % Smaller font
    \textbf{SR-B for mono-100 Models} \\[2pt] % Centered header with spacing    
    \tiny
    \begin{tabular}{lccc|ccc}
        \toprule
        \multirow{2}{*}{} & \multicolumn{3}{c|}{\textbf{Checkpoint 0}} & \multicolumn{3}{c}{\textbf{Checkpoint 4000}} \\
        & Random-mono-100 & \ofa-mono-100 & \frameworkname-mono 100 & Random-mono-100 & \ofa-mono-100 & \frameworkname-mono 100 \\
        \midrule
        eng\_Latn & - & - & - & - & - & - \\ 
        tur\_Latn & 5.2 & 5.6 & 6.4 & 8.2 & \textbf{11.2} & 9.4 \\
        ell\_Grek & 3.8 & 4.6 & 5.2 & 6.8 & \textbf{13.0} & 12.6 \\
        bul\_Cyrl & 4.8 & 5.6 & 3.8 & 13.8 & \textbf{29.2} & 28.8 \\
        ces\_Latn & 4.6 & 7.2 & 6.4 & 18.0 & 17.2 & \textbf{23.0} \\
        kor\_Hang & 3.6 & 6.0 & 6.4 & 7.8 & 10.6 & \textbf{12.0} \\
        kat\_Geor & 2.8 & 4.4 & 4.6 & 7.0 & 8.6 & \textbf{10.2} \\
        fry\_Latn & 3.6 & 5.6 & 7.2 & 14.6 & \textbf{16.8} & 15.4 \\
        zsm\_Latn & 3.8 & 6.6 & 6.6 & 11.8 & \textbf{23.4} & 20.2 \\
        khm\_Khmr & 2.8 & 5.8 & 4.6 & 3.8 & 6.2 & \textbf{8.0} \\
        jpn\_Japn & - & - & - & - & - & - \\ 
        yue\_Hani & 1.8 & 2.4 & 2.8 & 4.2 & \textbf{5.8} & \textbf{5.8} \\
        tuk\_Latn & 4.2 & 4.8 & \textbf{6.8} & 5.4 & 6.4 & 6.0 \\
        uig\_Arab & 2.2 & 3.2 & 3.2 & \textbf{4.0} & 3.8 & \textbf{4.0} \\
        pam\_Latn & 4.2 & 5.4 & 5.6 & 5.2 & 6.0 & \textbf{6.4} \\
        kab\_Latn & 2.8 & 2.4 & 3.6 & 3.8 & \textbf{5.2} & 4.2 \\
        gla\_Latn & 2.8 & 3.8 & \textbf{4.8} & 4.4 & 4.4 & 4.4 \\
        mhr\_Cyrl & 3.6 & 6.8 & \textbf{7.0} & 4.2 & 6.8 & 6.6 \\
        swh\_Latn & 3.4 & \textbf{5.0} & \textbf{5.0} & 3.8 & 4.8 & 3.6 \\
        cmn\_Hani & 5.8 & 5.2 & 3.8 & 5.0 & \textbf{9.0} & 8.0 \\
        pes\_Arab & 4.8 & \textbf{7.0} & 6.4 & 2.8 & 3.6 & 4.0 \\
        dtp\_Latn & 1.8 & 2.2 & 2.6 & \textbf{4.6} & 3.8 & \textbf{4.6} \\
        \bottomrule
    \end{tabular}
    \caption{Acc at 10 values in SR-B benchmark for Mono 100 models initialized with 3 approaches. Bold values highlight the best metric for each language.}
    \label{tab:lang_level_srb_results_mono100}
\end{table*}

\begin{table*}[h]
    \centering
    \footnotesize % Smaller font
    \textbf{SR-T for Mono 100 Models} \\[3pt] % Centered header with spacing    
    \tiny
    \begin{tabular}{lccc|ccc}
        \toprule
        \multirow{2}{*}{} & \multicolumn{3}{c|}{\textbf{Checkpoint 0}} & \multicolumn{3}{c}{\textbf{Checkpoint 4000}} \\
        & Random-mono-100 & \ofa-mono-100 & \frameworkname-mono 100 & Random-mono-100 & \ofa-mono-100 & \frameworkname-mono 100 \\
        \midrule
        eng\_Latn & - & - & - & - & - & - \\ 
        tur\_Latn & 3.2 & 4.2 & 5.2 & 9.4 & \textbf{15.6} & 8.6 \\ 
        ell\_Grek & 2.0 & 2.3 & 2.4 & 4.9 & \textbf{16.4} & 13.5 \\ 
        bul\_Cyrl & 3.7 & 4.3 & 4.4 & 20.8 & \textbf{48.5} & 42.1 \\ 
        ces\_Latn & 4.0 & 4.7 & 5.3 & 19.8 & \textbf{30.4} & 19.8 \\ 
        kor\_Hang & 2.8 & 4.4 & 4.1 & 7.4 & \textbf{11.3} & 8.3 \\ 
        kat\_Geor & 3.4 & 5.9 & 6.2 & 8.7 & \textbf{14.3} & 11.7 \\ 
        fry\_Latn & 19.7 & 23.7 & 27.8 & 40.5 & \textbf{46.8} & 35.3 \\ 
        zsm\_Latn & 5.2 & 9.8 & 9.6 & 13.9 & \textbf{34.1} & 22.3 \\ 
        khm\_Khmr & 2.6 & 4.6 & 4.3 & 3.9 & \textbf{9.8} & 6.9 \\ 
        jpn\_Japn & - & - & - & - & - & - \\ 
        yue\_Hani & 1.8 & 5.3 & 4.4 & 4.7 & \textbf{7.3} & 4.9 \\ 
        tuk\_Latn & 7.4 & 11.3 & 7.9 & 15.3 & \textbf{18.2} & 13.3 \\ 
        uig\_Arab & 2.1 & 2.3 & 2.4 & 2.3 & \textbf{2.6} & 2.0 \\ 
        pam\_Latn & 1.6 & 2.3 & 3.0 & 3.4 & \textbf{3.5} & 2.8 \\ 
        kab\_Latn & 2.0 & 2.2 & 2.9 & 2.9 & \textbf{3.4} & 2.4 \\ 
        gla\_Latn & 3.4 & 4.5 & 4.1 & 4.1 & \textbf{4.7} & 4.2 \\ 
        mhr\_Cyrl & 2.4 & 3.2 & 2.5 & 2.8 & \textbf{4.1} & 3.5 \\ 
        swh\_Latn & 11.3 & 11.5 & 11.3 & 13.1 & \textbf{15.6} & 11.3 \\ 
        cmn\_Hani & 3.7 & 4.8 & 3.7 & 4.9 & \textbf{9.4} & 7.5 \\ 
        pes\_Arab & 2.9 & 4.2 & \textbf{4.3} & 2.6 & 2.9 & 2.1 \\ 
        dtp\_Latn & 3.1 & 3.3 & 4.0 & 3.9 & \textbf{5.1} & 3.5 \\ 
        \bottomrule
    \end{tabular}
    \caption{Acc at 10 values in SR-T benchmark for Mono 100 models initialized with 3 approaches. Bold values highlight the best metric for each language.}
    \label{tab:lang_level_srt_results_mono100}
\end{table*}

\begin{table*}[h]
    \centering
    \footnotesize % Smaller font
    \textbf{NER for Mono 100 Models} \\[3pt] % Centered header with spacing    
    \tiny
    \begin{tabular}{lccc|ccc}
        \toprule
        \multirow{2}{*}{} & \multicolumn{3}{c|}{\textbf{Checkpoint 0}} & \multicolumn{3}{c}{\textbf{Checkpoint 4000}} \\
        & Random-mono-100 & \ofa-mono-100 & \frameworkname-mono 100 & Random-mono-100 & \ofa-mono-100 & \frameworkname-mono 100 \\
        \midrule
        eng\_Latn & 75.9 & 75.3 & 75.4 & \textbf{80.9} & 80.5 & 80.6 \\ 
        tur\_Latn & 32.0 & 32.8 & 32.3 & 47.7 & \textbf{55.9} & 52.1 \\ 
        ell\_Grek & 10.7 & 10.2 & 9.8 & 37.0 & \textbf{47.2} & 45.0 \\ 
        bul\_Cyrl & 19.0 & 20.5 & 24.0 & 54.3 & 64.7 & \textbf{65.5} \\ 
        ces\_Latn & 36.1 & 37.8 & 37.4 & 59.6 & \textbf{61.9} & 61.4 \\ 
        kor\_Hang & 11.3 & 13.8 & 10.9 & 17.1 & \textbf{29.2} & 27.3 \\ 
        kat\_Geor & 11.9 & 14.6 & 14.0 & 25.9 & \textbf{34.8} & 30.9 \\ 
        fry\_Latn & 29.9 & 30.2 & 32.0 & 68.0 & \textbf{70.1} & 65.9 \\ 
        zsm\_Latn & - & - & - & - & - & - \\ 
        khm\_Khmr & 17.2 & 17.4 & 14.6 & 30.7 & \textbf{35.9} & 32.6 \\ 
        jpn\_Japn & - & - & - & - & - & - \\ 
        yue\_Hani & 7.7 & 7.4 & 6.0 & 9.2 & \textbf{14.3} & 12.1 \\ 
        tuk\_Latn & 24.4 & 25.2 & 26.9 & \textbf{41.7} & 40.6 & 40.0 \\ 
        uig\_Arab & 14.7 & 14.6 & 16.9 & \textbf{20.9} & 16.4 & 18.7 \\ 
        pam\_Latn & - & - & - & - & - & - \\ 
        kab\_Latn & - & - & - & - & - & - \\ 
        gla\_Latn & 25.7 & 24.9 & 20.3 & 45.0 & \textbf{51.5} & 39.3 \\ 
        mhr\_Cyrl & 9.4 & 11.1 & 8.6 & 21.6 & \textbf{36.2} & 36.1 \\ 
        swh\_Latn & - & - & - & - & - & - \\ 
        cmn\_Hani & - & - & - & - & - & - \\ 
        pes\_Arab & - & - & - & - & - & - \\ 
        dtp\_Latn & - & - & - & - & - & - \\ 
        \bottomrule
    \end{tabular}
    \caption{F1 scores in NER benchmark for Mono 100 models. Bold values highlight the best metric for the language.}
    \label{tab:lang_level_ner_results_mono100} % Adjusted label
\end{table*}

\begin{table*}[h]
    \centering
    \footnotesize % Smaller font
    \textbf{POS for Mono 100 Models} \\[3pt] % Centered header with spacing    
    \tiny
    \begin{tabular}{lccc|ccc}
        \toprule
        \multirow{2}{*}{} & \multicolumn{3}{c|}{\textbf{Checkpoint 0}} & \multicolumn{3}{c}{\textbf{Checkpoint 4000}} \\
        & Random-mono-100 & \ofa-mono-100 & \frameworkname-mono 100 & Random-mono-100 & \ofa-mono-100 & \frameworkname-mono 100 \\
        \midrule
        eng\_Latn & 94.8 & 94.9 & 94.9 & \textbf{95.8} & \textbf{95.8} & \textbf{95.8} \\ 
        tur\_Latn & 25.7 & 26.9 & 26.5 & 41.9 & 48.4 & \textbf{49.2} \\ 
        ell\_Grek & 16.8 & 18.3 & 17.2 & 54.0 & 75.3 & \textbf{76.5} \\ 
        bul\_Cyrl & 21.8 & 24.2 & 23.1 & 77.8 & 82.8 & \textbf{83.7} \\ 
        ces\_Latn & 25.3 & 27.0 & 26.2 & 78.0 & 79.4 & \textbf{80.4} \\ 
        kor\_Hang & 19.9 & 21.9 & 20.9 & 35.6 & \textbf{40.7} & 40.1 \\ 
        kat\_Geor & - & - & - & - & - & - \\ 
        fry\_Latn & - & - & - & - & - & - \\ 
        zsm\_Latn & - & - & - & - & - & - \\ 
        khm\_Khmr & 20.9 & 20.3 & 23.1 & 13.2 & \textbf{15.3} & 10.4 \\ 
        jpn\_Japn & - & - & - & - & - & - \\
        yue\_Hani & 16.8 & 17.5 & 17.5 & \textbf{32.7} & 32.6 & 30.7 \\ 
        tuk\_Latn & - & - & - & - & - & - \\ 
        uig\_Arab & - & - & - & - & - & - \\ 
        pam\_Latn & - & - & - & - & - & - \\ 
        kab\_Latn & 20.2 & 20.9 & 20.8 & 31.1 & \textbf{40.7} & 39.8 \\ 
        gla\_Latn & - & - & - & - & - & - \\ 
        mhr\_Cyrl & - & - & - & - & - & - \\ 
        swh\_Latn & - & - & - & - & - & - \\ 
        cmn\_Hani & - & - & - & - & - & - \\ 
        pes\_Arab & - & - & - & - & - & - \\ 
        dtp\_Latn & - & - & - & - & - & - \\ 
        \bottomrule
    \end{tabular}
    \caption{F1 scores in POS benchmark for Mono 100 models. Bold values highlight the best metric for the language.}
    \label{tab:lang_level_pos_results_mono100}
\end{table*}

\begin{table*}[h]
    \centering
    \footnotesize
    \textbf{SR-B for Multi 400 Models} \\[2pt]
    \tiny
    \begin{tabular}{lccc|ccc}
        \toprule
        \multirow{2}{*}{} & \multicolumn{3}{c|}{\textbf{Checkpoint 0}} & \multicolumn{3}{c}{\textbf{Checkpoint 4000}} \\
        & Random-multi-400 & \ofa-multi-400 & \frameworkname-multi 400 & Random-multi-400 & \ofa-multi-400 & \frameworkname-multi 400 \\
        \midrule
        eng\_Latn & - & - & - & - & - & - \\
        tur\_Latn & 13.6 & 13.6 & 15.8 & 75.4 & \textbf{76.0} & \textbf{76.0} \\
        ell\_Grek & 6.2 & 6.6 & 8.2 & 50.0 & 49.6 & \textbf{50.8} \\
        bul\_Cyrl & 16.6 & 15.4 & 15.6 & \textbf{82.4} & 82.0 & \textbf{82.4} \\
        ces\_Latn & 15.8 & 18.4 & 17.8 & 73.4 & \textbf{74.6} & 74.4 \\
        kor\_Hang & 9.8 & 9.8 & 9.8 & \textbf{63.2} & 62.4 & 62.8 \\
        kat\_Geor & 3.0 & 4.8 & 6.2 & 43.2 & \textbf{44.2} & 43.8 \\
        fry\_Latn & 5.0 & 5.6 & 5.6 & 49.0 & 50.0 & \textbf{51.0} \\
        zsm\_Latn & 17.2 & 18.2 & 18.6 & 80.4 & \textbf{84.6} & 84.4 \\
        khm\_Khmr & 3.6 & 3.0 & 4.2 & 30.8 & \textbf{31.6} & 31.4 \\
        jpn\_Japn & - & - & - & - & - & - \\
        yue\_Hani & 3.0 & 3.2 & 3.4 & \textbf{13.6} & 13.0 & 12.8 \\
        tuk\_Latn & 5.6 & 4.4 & 5.4 & 46.0 & 54.4 & \textbf{54.6} \\
        uig\_Arab & 4.6 & 7.0 & 6.6 & 33.8 & \textbf{34.8} & 34.6 \\
        pam\_Latn & 5.2 & 4.2 & 4.4 & 20.4 & 21.0 & \textbf{23.2} \\
        kab\_Latn & 3.0 & 4.0 & 3.0 & 8.0 & \textbf{10.4} & 9.4 \\
        gla\_Latn & 4.0 & 3.6 & 4.0 & \textbf{28.6} & 27.4 & 25.8 \\
        mhr\_Cyrl & 3.2 & 3.8 & 3.6 & 20.0 & 25.0 & \textbf{25.2} \\
        swh\_Latn & 8.2 & 9.6 & 8.6 & 34.8 & \textbf{40.0} & 38.0 \\
        cmn\_Hani & 17.4 & 17.8 & 17.2 & 28.2 & \textbf{30.0} & 28.2 \\
        pes\_Arab & 14.2 & 16.4 & 22.2 & 28.6 & \textbf{30.4} & 29.6 \\
        dtp\_Latn & 2.6 & 2.6 & 3.4 & \textbf{5.2} & \textbf{5.2} & 4.6 \\
        \bottomrule
    \end{tabular}
    \caption{Acc@10 values in SR-B benchmark for Multi 400 models initialized with 3 approaches. Bold values highlight the best metric for each language.}
    \label{tab:lang_level_srb_results_multi400}
\end{table*}

\begin{table*}[h]
    \centering
    \footnotesize
    \textbf{SR-T for Multi 400 Models} \\[2pt]
    \tiny
    \begin{tabular}{lccc|ccc}
        \toprule
        \multirow{2}{*}{} & \multicolumn{3}{c|}{\textbf{Checkpoint 0}} & \multicolumn{3}{c}{\textbf{Checkpoint 4000}} \\
        & Random-multi-400 & \ofa-multi-400 & \frameworkname-multi 400 & Random-multi-400 & \ofa-multi-400 & \frameworkname-multi 400 \\
        \midrule
        eng\_Latn & - & - & - & - & - & - \\
        tur\_Latn & 22.8 & 22.2 & 23.0 & 87.7 & \textbf{87.8} & 87.4 \\
        ell\_Grek & 21.2 & 21.0 & 20.3 & 79.8 & \textbf{80.8} & 80.2 \\
        bul\_Cyrl & 34.4 & 35.7 & 36.1 & \textbf{88.3} & 88.1 & 88.2 \\
        ces\_Latn & 25.3 & 25.3 & 25.5 & 83.2 & \textbf{84.6} & 83.4 \\
        kor\_Hang & 21.1 & 21.3 & 21.4 & \textbf{79.3} & 79.1 & 78.9 \\
        kat\_Geor & 12.1 & 13.1 & 12.1 & 63.5 & \textbf{64.6} & \textbf{64.6} \\
        fry\_Latn & 35.3 & 33.5 & 33.0 & 84.4 & \textbf{86.7} & 83.8 \\
        zsm\_Latn & 31.4 & 32.2 & 32.7 & 90.5 & \textbf{91.4} & 90.7 \\
        khm\_Khmr & 5.0 & 4.6 & 5.3 & 51.8 & \textbf{52.6} & 52.4 \\
        jpn\_Japn & - & - & - & - & - & - \\
        yue\_Hani & 22.1 & 22.5 & 22.3 & 63.8 & 59.4 & \textbf{64.9} \\
        tuk\_Latn & 14.3 & 15.3 & 14.3 & 48.8 & \textbf{51.2} & \textbf{51.2} \\
        uig\_Arab & 7.0 & 8.0 & 7.8 & 54.2 & 56.3 & \textbf{57.2} \\
        pam\_Latn & 4.4 & 4.8 & 4.5 & 7.0 & \textbf{7.8} & 7.5 \\
        kab\_Latn & 2.5 & 3.6 & 3.1 & 7.9 & 7.4 & \textbf{8.9} \\
        gla\_Latn & 5.4 & 5.3 & 5.3 & 33.2 & \textbf{36.1} & 33.2 \\
        mhr\_Cyrl & 2.8 & 3.2 & 3.6 & 17.8 & 20.2 & \textbf{22.5} \\
        swh\_Latn & 21.0 & 20.5 & 20.5 & 35.4 & \textbf{36.4} & 36.2 \\
        cmn\_Hani & 33.1 & 33.5 & 32.9 & \textbf{65.0} & 60.7 & 62.5 \\
        pes\_Arab & 27.2 & 28.5 & 27.6 & 59.3 & 57.4 & \textbf{63.1} \\
        dtp\_Latn & 3.6 & 4.1 & 3.5 & 5.7 & \textbf{6.3} & 5.4 \\
        \bottomrule
    \end{tabular}
    \caption{Acc@10 values in SR-T benchmark for Multi 400 models initialized with 3 approaches. Bold values highlight the best metric for each language.}
    \label{tab:lang_level_srt_results_multi400}
\end{table*}

\begin{table*}[h]
    \centering
    \footnotesize
    \textbf{NER for Multi 400 Models} \\[2pt]
    \tiny
    \begin{tabular}{lccc|ccc}
        \toprule
        \multirow{2}{*}{} & \multicolumn{3}{c|}{\textbf{Checkpoint 0}} & \multicolumn{3}{c}{\textbf{Checkpoint 4000}} \\
        & Random-multi-400 & \ofa-multi-400 & \frameworkname-multi 400 & Random-multi-400 & \ofa-multi-400 & \frameworkname-multi 400 \\
        \midrule
        eng\_Latn & 78.1 & 78.4 & 77.9 & \textbf{81.3} & 81.2 & \textbf{81.3} \\
        tur\_Latn & 55.9 & 59.2 & 58.3 & \textbf{72.8} & \textbf{72.8} & 72.0 \\
        ell\_Grek & 58.1 & 56.8 & 59.6 & \textbf{70.6} & 69.3 & 70.2 \\
        bul\_Cyrl & 63.7 & 64.4 & 64.3 & \textbf{76.9} & 76.4 & 76.0 \\
        ces\_Latn & 61.7 & 61.2 & 61.5 & 75.9 & 75.8 & \textbf{76.0} \\
        kor\_Hang & 39.8 & 41.2 & 41.1 & 48.5 & 48.8 & \textbf{49.1} \\
        kat\_Geor & 48.9 & 52.1 & 53.1 & 62.2 & 62.3 & \textbf{62.9} \\
        fry\_Latn & 56.3 & 58.8 & 56.6 & 78.1 & \textbf{78.4} & 76.9 \\
        zsm\_Latn & - & - & - & - & - & - \\
        khm\_Khmr & 36.1 & 37.3 & 33.5 & 45.2 & 43.5 & \textbf{47.1} \\
        jpn\_Japn & - & - & - & - & - & - \\
        yue\_Hani & 20.7 & 20.0 & 23.8 & 16.2 & \textbf{23.8} & 21.0 \\
        tuk\_Latn & 30.2 & 34.3 & 35.5 & 56.7 & \textbf{57.9} & 55.7 \\
        uig\_Arab & 28.2 & 34.6 & 34.8 & \textbf{48.2} & 47.0 & 45.5 \\
        pam\_Latn & - & - & - & - & - & - \\
        kab\_Latn & - & - & - & - & - & - \\
        gla\_Latn & 37.5 & 39.2 & 38.0 & 56.8 & 55.9 & \textbf{61.7} \\
        mhr\_Cyrl & 27.8 & 23.7 & 27.5 & 48.3 & 51.0 & \textbf{51.1} \\
        swh\_Latn & - & - & - & - & - & - \\
        cmn\_Hani & - & - & - & - & - & - \\
        pes\_Arab & - & - & - & - & - & - \\
        dtp\_Latn & - & - & - & - & - & - \\
        \bottomrule
    \end{tabular}
    \caption{F1 scores in NER benchmark for Multi 400 models initialized with 3 approaches. Bold values highlight the best metric for each language.}
    \label{tab:lang_level_ner_results_multi400}
\end{table*}

\begin{table*}[h]
    \centering
    \footnotesize % Smaller font
    \textbf{POS for Multi 400 Models} \\[3pt] % Centered header with spacing    
    \tiny
    \begin{tabular}{lccc|ccc}
        \toprule
        \multirow{2}{*}{} & \multicolumn{3}{c|}{\textbf{Checkpoint 0}} & \multicolumn{3}{c}{\textbf{Checkpoint 4000}} \\
        & Random-multi-400 & \ofa-multi-400 & \frameworkname-multi 400 & Random-multi-400 & \ofa-multi-400 & \frameworkname-multi 400 \\
        \midrule
        eng\_Latn & 95.3 & 95.4 & 95.3 & \textbf{95.8} & \textbf{95.8} & \textbf{95.8} \\ 
        tur\_Latn & 62.4 & 61.5 & 62.4 & \textbf{71.4} & \textbf{71.4} & 71.3 \\ 
        ell\_Grek & 84.6 & 83.4 & 84.0 & \textbf{86.0} & 85.9 & \textbf{86.0} \\ 
        bul\_Cyrl & 85.9 & 85.6 & 86.1 & 87.8 & \textbf{88.0} & \textbf{88.0} \\ 
        ces\_Latn & 74.3 & 73.9 & 73.0 & \textbf{82.7} & \textbf{82.7} & 82.5 \\ 
        kor\_Hang & 52.0 & 52.2 & 52.5 & 52.4 & \textbf{52.6} & 52.5 \\ 
        kat\_Geor & - & - & - & - & - & - \\ 
        fry\_Latn & - & - & - & - & - & - \\ 
        zsm\_Latn & - & - & - & - & - & - \\ 
        khm\_Khmr & - & - & - & - & - & - \\ 
        jpn\_Japn & - & - & - & - & - & - \\ 
        yue\_Hani & 40.2 & 25.5 & 28.6 & 27.2 & \textbf{27.3} & 27.1 \\ 
        tuk\_Latn & - & - & - & - & - & - \\ 
        uig\_Arab & 58.8 & 57.5 & 57.9 & \textbf{69.2} & 68.9 & 68.9 \\ 
        pam\_Latn & - & - & - & - & - & - \\ 
        kab\_Latn & - & - & - & - & - & - \\ 
        gla\_Latn & 31.7 & 31.4 & 33.6 & 60.4 & \textbf{60.7} & 60.6 \\ 
        mhr\_Cyrl & - & - & - & - & - & - \\ 
        swh\_Latn & - & - & - & - & - & - \\ 
        cmn\_Hani & - & - & - & - & - & - \\ 
        pes\_Arab & - & - & - & - & - & - \\ 
        dtp\_Latn & - & - & - & - & - & - \\ 
        \bottomrule
    \end{tabular}
    \caption{F1 scores in POS benchmark for Multi 400 models initialized with 3 approaches. Bold values highlight the best metric for the language.}
    \label{tab:lang_level_pos_results_multi400}
\end{table*}

%%%%%%%%%%%%%%%%%%%%%%%%%%%%%%%%%%%%%%%%%%%%%%%%%%%%%%%%%%%%

\end{document}